\documentclass[11pt]{article}

\usepackage[final]{acl}

\usepackage{times}
\usepackage{latexsym}
\usepackage[T1]{fontenc}
\usepackage[utf8]{inputenc}
\usepackage{microtype}
\usepackage{inconsolata}
\usepackage{graphicx}
\usepackage{xcolor}
\usepackage{tabularx}
\usepackage{booktabs}
\usepackage{amsmath}
\usepackage{algorithm}
\usepackage{algpseudocode}
\usepackage{listings}

\definecolor{promptbg}{gray}{0.97}
\definecolor{promptred}{rgb}{0.6,0.0,0.0}
\lstdefinestyle{prompt}{
  basicstyle=\ttfamily\footnotesize\color{promptred},
  breaklines=true,
  breakatwhitespace=false,
  columns=fullflexible,
  keepspaces=true,
  frame=single,
  rulecolor=\color{black!30},
  framesep=4pt,
  backgroundcolor=\color{promptbg},
  xleftmargin=5pt,
  xrightmargin=3pt,
  aboveskip=6pt,
  belowskip=4pt,
  showstringspaces=false,
}
\newcommand{\promptlabel}[1]{\vspace{2pt}\noindent\textbf{\small #1}\par\nobreak\vspace{1pt}}

\title{Litmus: Zero-Label, Code-Driven Metric Specification for Evaluating AI Systems}

\author{
  \textbf{Prajjwal Gupta},
  \textbf{Prasang Gupta},
  \textbf{Vishal Bhutani},
  \textbf{Apoorva Sharma},
\\
  \textbf{Sumanth Chundru},
  \textbf{Waqar Sarguroh},
  \textbf{Kevin Paul}
\\
\\
  PricewaterhouseCoopers, U.S.
}

\begin{document}
\maketitle

\begin{abstract}
As agentic LLM systems move from prototypes to deployment, evaluating them has become more important and more difficult. Evaluation goals are often left implicit. Without a clear account of what a system is expected to do, how it can fail, and which failures matter, metric choices become difficult to justify, interpret, or validate. We present Litmus, a zero-label system that designs evaluation and monitoring metrics for AI pipelines by eliciting evaluation intent from source code and targeted interrogation. Litmus first identifies what must be measured and why, then converts those answers into constraints for a justified, per-stage metric portfolio. We evaluate Litmus on three real, code-defined pipelines against AutoMetrics and three DynamicRubric baselines. Litmus achieves broader concern coverage on account grouping at matched size, spans more stages, produces a near-zero-redundancy portfolio, and, under a zero-label design-time constraint, ranks first in validity against per-row quality labels on all three pipelines, decisively on scientific QA ($\rho{=}0.72$ vs.\ $0.47$ for the strongest baseline on the stitched single-vector score), with overlapping intervals on the other two. Results support a shift from automatic metric implementation to automatic metric specification: before asking which metric to compute, evaluation systems should ask what must be measured and why.
\end{abstract}

\begin{figure}[t!]
  \includegraphics[width=\columnwidth]{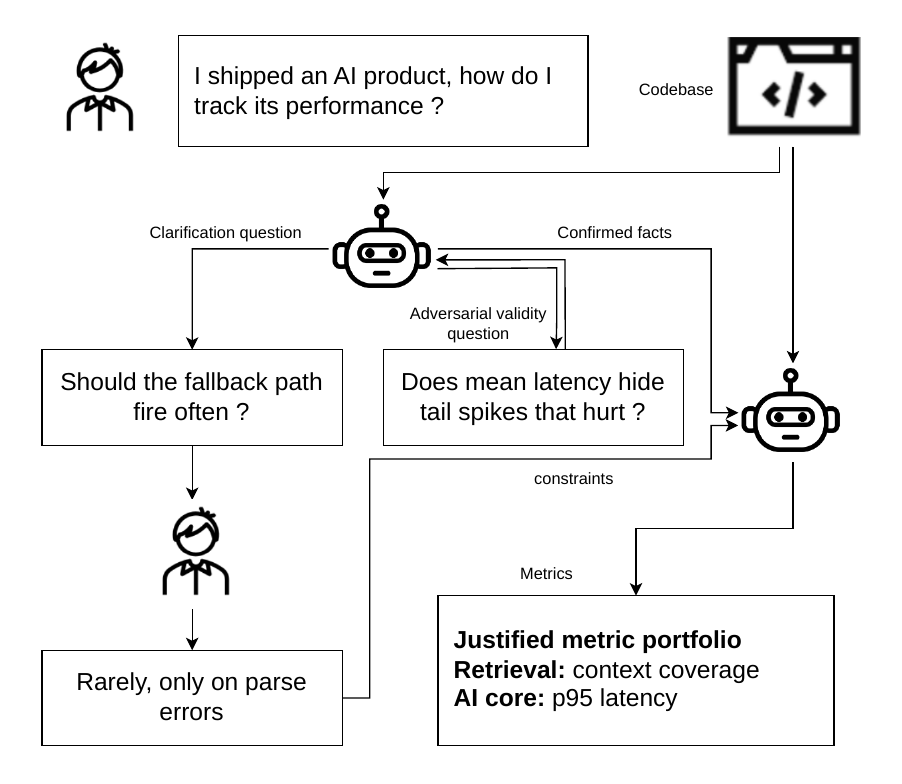}
  \caption{Litmus designs metrics by asking questions first. From source code and a practitioner's goal alone, it interrogates the pipeline through practitioner-facing clarification questions and internal adversarial validity questions. The answers become facts that act as constraints, yielding a justified, per-stage metric portfolio.}
  \label{fig:teaser}
\end{figure}

\section{Introduction}

LLM-based systems are increasingly moving from demonstrations to production workflows in domains such as finance, customer support, software engineering, healthcare administration, and enterprise knowledge management. Many of these systems are no longer single prompt-response models: they are agentic or pipeline-based applications that combine retrieval, tool use, routing logic, structured outputs, business rules, fallbacks, confidence estimates, and human review. This shift changes the role of evaluation. In research settings, evaluation is often used to compare model outputs; in production settings, it must also support debugging, monitoring, auditability, and risk management.

This distinction matters because production failures are rarely explained by a single final-output score. A financial-document pipeline may fail because retrieval returned the wrong context, because a rule-based override was skipped, because a fallback path fired too often, because confidence was miscalibrated, or because a downstream formatter masked an upstream error. Conversely, a behavior that appears undesirable in isolation may be correct under the system design: frequent fallbacks may indicate model weakness in one deployment but appropriate caution in another. For industry practitioners, the practical question is therefore not only whether an output is good, but which component produced the behavior, whether that behavior was expected, and which operational risk should be monitored.

Current evaluation practice is poorly matched to this need. Automatic metrics have become the default instrument for evaluating natural language generation (NLG) systems: a recent survey of 110 ACL and INLG papers finds that 94\% report at least one automatic metric \citep{schmidtova-etal-2024-automatic-metrics}. Yet the same survey finds that 76.9\% of reported metric usages provide no rationale, and that unclear evaluation goals contribute to a ``kitchen-sink'' style of reporting many weakly motivated scores \citep{schmidtova-etal-2024-automatic-metrics,zhou-etal-2022-deconstructing}. This is downstream of a more basic issue: a metric cannot be justified, interpreted, operationalized, or monitored against an evaluation goal that has never been made explicit.

For LLM pipelines, the practical process of translating source-code structure and practitioner intent into concrete evaluation and monitoring metrics remains under-specified. We argue that metric design should begin with an elicitation step that makes the evaluation goal explicit. The question that precedes \emph{which metric?} is: \emph{what is this component supposed to do, how can it fail silently, and what is therefore worth measuring?} This is fundamentally an inquiry problem---the same uncertainty-resolution view that underlies clarification-question generation \citep{rao-daume-iii-2018-learning}. We adopt this view for AI-system evaluation. Rather than treating metrics as generic objects to be selected after the fact, we treat them as measurement commitments whose validity depends on unresolved facts about system design, deployment intent, and failure semantics.

We instantiate this view in \textbf{Litmus} (Figure~\ref{fig:teaser}), a zero-label system that designs evaluation and monitoring metrics for an AI pipeline directly from its source code. Litmus first builds a code-grounded model of the pipeline: its major components, their roles, the AI capabilities they use, and the failure surfaces they expose. It then maps these components to an evaluation-pattern taxonomy to identify plausible metric families. However, code can expose where measurement may be needed without fully specifying what should count as success, failure, or acceptable risk. Litmus therefore interrogates the pipeline to convert such ambiguities into metric-design constraints. Practitioner-facing clarification questions target under-specified decisions that affect metric design, such as which processing tier carries production traffic or whether a fallback path is expected to fire often or rarely. Internal adversarial validity questions challenge each candidate metric on pipeline fit, data assumptions, measurement validity, and direction of goodness. The resulting answers become \emph{confirmed facts}: constraints that admit, reject, or re-scope metrics. Litmus therefore outputs not a single holistic score, but a justified portfolio of stage-specific evaluation and monitoring metrics and is currently deployed and being used to support governance of AI tooling across multiple client projects. This establishes operational use; it does not establish improvement over manual or heuristic metric design.

Our contributions are threefold:
\begin{enumerate}
  \item We reframe automatic metric design as \emph{goal elicitation by interrogation}, connecting metric validity concerns in NLG evaluation to clarification-question generation and inquiry-based uncertainty resolution.
  \item We introduce \textbf{Litmus} (Figure~\ref{fig:pipeline}), a zero-label system that derives a justified, per-stage evaluation and monitoring portfolio from source code, with practitioner answers and internal validity checks represented as explicit metric-design constraints.
  \item We evaluate Litmus on three real, code-defined AI pipelines spanning distinct domains---financial account grouping, scientific QA, and inherent risk assessment---against AutoMetrics and DynamicRubric baselines. To assess metric-design quality beyond label agreement, we additionally introduce portfolio-level axes---coverage, grounding, and redundancy---that characterize what an automatic metric-design system produces.
\end{enumerate}

\section{Related Work}
\label{sec:related}

Evaluation in deployed AI systems serves a broader role than benchmark comparison: it must support debugging, regression testing, monitoring, auditability, and operational decision-making. Prior work on production machine learning emphasizes that deployed ML systems accumulate hidden technical debt through data dependencies, configuration assumptions, feedback loops, and boundary failures \citep{NIPS2015_86df7dcf}. \citet{8258038} argue for systematic tests that go beyond aggregate model quality, including data validation, infrastructure checks, and monitoring; \citet{8804457} similarly show that engineering AI-enabled systems requires connecting model behavior to data, code, and deployment context. Documentation and auditing frameworks make a related point: intended use, limitations, assumptions, and evaluation evidence should be explicit in deployed AI systems \citep{10.1145/3287560.3287596,10.1145/3458723,10.1145/3351095.3372873}. Litmus follows this production-oriented view: it treats metric design as a system-level task in which metrics should be grounded in components, failure surfaces, and monitoring needs rather than only in final outputs.

A growing body of work scrutinizes how metrics are used and reported. \citet{schmidtova-etal-2024-automatic-metrics} survey current NLG evaluation practice and find pervasive missing rationales, missing implementation details, and missing correlations with human judgement; \citet{zhou-etal-2022-deconstructing} trace the ``kitchen-sink'' tendency to unclear evaluation goals. Validity critiques of overlap metrics \citep{reiter-2018-structured,papineni-etal-2002-bleu,lin-2004-rouge} motivate the move toward LLM-as-judge evaluation \citep{liu-etal-2023-g,fu-etal-2024-gptscore}. However, LLM-as-judge methods do not remove the need for specification: a judge prompt still encodes assumptions about what matters, what evidence should be considered, and how scores should be interpreted. Prior work also shows that LLM judges may exhibit systematic biases and model-family alignment effects \citep{zheng2023judging}. Litmus targets the specification gap these works identify: it produces metrics whose rationale is a first-class artifact.

Recent benchmarks measure general agent capability across reasoning-and-acting, tool use, web navigation, and software-engineering tasks \citep{yao2023react,qin2024toolllm,liu2024agentbench,zhou2024webarena,jimenez2024swebench}. Deployed industry pipelines, however, contain domain-specific rules, retrieval layers, routing logic, fallbacks, confidence thresholds, and compliance constraints. Litmus addresses a complementary problem: designing metrics for a particular code-defined pipeline by identifying what should be measured, at which stage, and why.

The closest conceptual neighbour to our framing is clarification-question generation. \citet{rao-daume-iii-2018-learning} formalize ``a good question is one whose expected answer is most useful'' via the expected value of perfect information, \citet{rao-daume-iii-2019-answer} generate rather than merely rank such questions, and \citet{majumder-etal-2021-ask} identify information ``essential to accomplish an underlying goal but currently missing from the context.'' Litmus adopts this usefulness-of-answer stance: every question it surfaces is annotated with the specific metric decision its answer would resolve. In our setting, the missing information is not needed to answer a user query, but to determine whether a candidate metric is appropriate, what it should measure, and how it should be interpreted. We distinguish this from mainstream question generation (QG/NQG), which generates answerable comprehension questions from a passage \citep{mulla2023automatic,ijcai2024p889,Flor2025}---a different speech act from resolving evaluation-intent ambiguity.

Recent systems automate parts of evaluation construction. \emph{AutoMetrics} takes outputs plus labels and fits a single runnable holistic judge -- a metric implementation system \citep{ryan2026autometrics}. \emph{DynamicRubric} \citep{wang2026generatingrefiningdynamicevaluation} has the judge generate its own rubric and then score, in per-instance (\textsc{-Inst}) and dataset-wide (\textsc{-DS}) variants, with a DPO-fine-tuned generator variant (\textsc{-FineTuned}). \emph{AutoRubric} \citep{jia-etal-2026-autorubric} builds a rubric for a given target, and \emph{EvalAssist} \citep{desmond2025evalassist} helps a user interactively refine judging criteria once chosen. All assume the evaluation target is given and focus on the final output. Litmus instead elicits the target first and designs a per-stage portfolio from source code with no labels, determining what to measure, where, and why before any judge is written. This distinction is especially important for production AI pipelines, where failures may arise in intermediate components and where monitoring requires metrics that are traceable to system behavior.

\section{The Litmus System}
\label{sec:litmus-system}

\begin{figure*}[t]
  \centering
  \includegraphics[width=\textwidth]{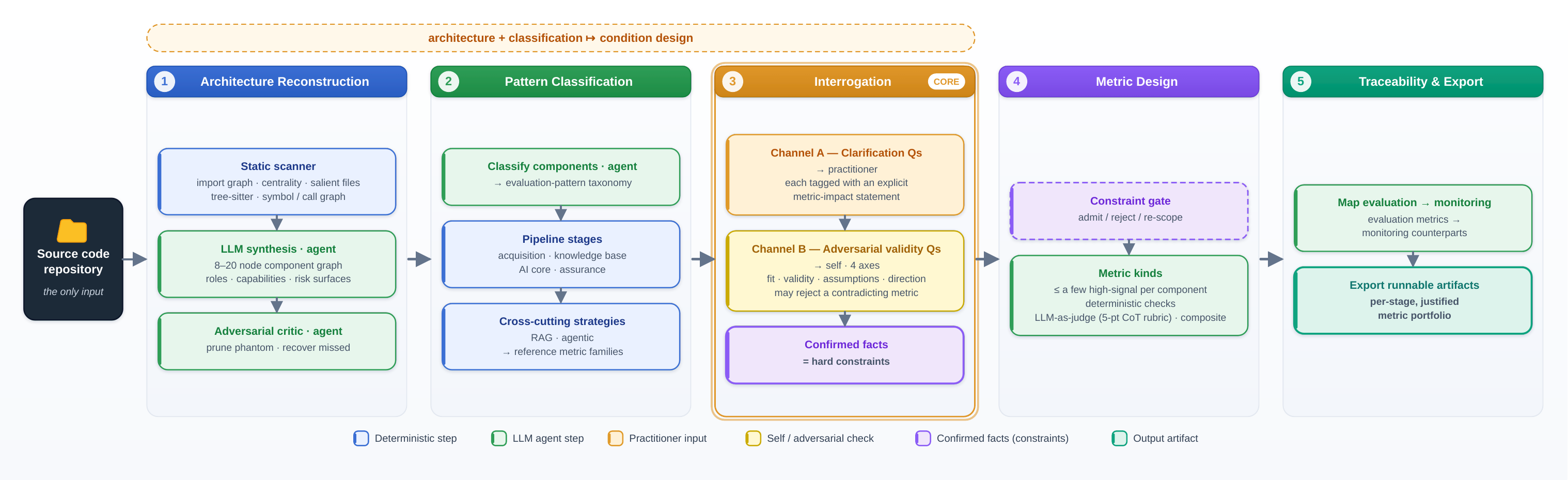}
  \caption{The Litmus pipeline. Static analysis and LLM synthesis reconstruct a validated component architecture (Phases~1--2); the system then interrogates the artifact and the practitioner (Phase~3), and the resulting \emph{confirmed facts} become hard constraints that admit, reject, or re-scope each candidate metric (Phase~4) before export (Phase~5).
  }
  \label{fig:pipeline}
\end{figure*}

Because code structure alone cannot determine intended behavior, acceptable failure modes, the available evidence for measurement, or even the direction in which a metric should improve, Litmus treats metric design as a constrained synthesis problem, organized into the stages below (Figure~\ref{fig:pipeline}).

\subsection{Architecture Reconstruction}

Litmus begins by constructing an evidence-grounded representation of the repository. A deterministic scanner parses the codebase, builds import and symbol/call graphs, computes centrality scores, and selects salient files for LLM analysis. An LLM then synthesizes an 8--20 node component graph describing the system's major modules, their roles, AI capabilities, and associated risk surfaces, and an adversarial critic checks it against code evidence, pruning phantom components and flagging missed ones, to reduce over-interpretation. The result is a repository-level architecture that is both structural (rooted in dependency and call relationships) and semantic (components are labeled with their functional roles); full prompts are given in Appendix~\ref{app:impl-arch}, with architecture-reconstruction detail in Appendix~\ref{app:arch}.

\subsection{From Patterns to Constraints}

Litmus next maps each component to an evaluation taxonomy of pipeline-stage patterns (data acquisition, knowledge base, AI core, assurance) and cross-cutting strategies (e.g., retrieval-augmented generation, agentic orchestration). This does not select final metrics; it narrows the design space by flagging which metric families are plausibly relevant per component, e.g., a retrieval component activates coverage, grounding, and latency families, while leaving the underlying assumptions unresolved. Litmus settles those assumptions through interrogation: rather than silently assuming answers, it converts each metric-affecting uncertainty into an explicit question whose answer is recorded as a confirmed fact: a user- or critic-validated constraint that downstream metric design must satisfy. Confirmed facts may determine whether a metric is included, how it is computed and thresholded, what data must be instrumented, how multiple signals are aggregated, or which direction indicates improvement. For example, the clarification ``should an abstention (`insufficient evidence') count as success or failure?'' fixes the \emph{direction of goodness} of the abstention-rate metric---the two answers invert its production alert.

\subsection{Metric Design and Export}

Only after interrogation does Litmus finalize the portfolio. Conditioned on the reconstructed architecture, the component classifications, and the confirmed facts, the designer emits a small number of high-signal metrics per component: deterministic checks, LLM-as-judge metrics reserved for semantic quality (a five-level rubric and an evidence-backed rationale), and composite metrics; each specifying its evaluated component, purpose, required data, computation, direction of goodness, and validity conditions. The result is a justified specification rather than a list of names: every metric is linked to the evidence and confirmed facts that support it, so the portfolio can be audited. Re-scanning a changed repository regenerates only the metrics whose supporting evidence or confirmed facts have moved. Litmus exports runnable definitions with this traceability and, where possible, maps each evaluation metric to a production monitoring counterpart, carrying the same rationale from offline design to deployment.

\section{Experimental Setup}
\label{sec:setup}

\subsection{Evaluation Domains}
\label{sec:pipelines}

Because Litmus is a source-code-based metric-design system while the baselines construct final-output judges or rubrics, we interpret the results as trade-offs under a zero-label design-time constraint. We evaluated three real, code-defined AI pipelines from different domains, testing whether the benefits of eliciting evaluation targets generalize across production-style systems with different tasks, data, and failure modes rather than being specific to one artifact.

\paragraph{Account grouping} This audit task classifies individual client general-ledger accounts into standardized, aggregated financial-statement categories. For each account the pipeline emits an \texttt{account\_group}, a confidence score, and a natural-language reason, via multiple decision paths (rule-based overrides, disaggregation logic, and retrieval-augmented mapping).

\paragraph{Scientific QA} We use the established PaperQA2/LitQA2 \citep{skarlinski2024languageagentsachievesuperhuman} question-answering setting and evaluate the answer-synthesis stage on a key-passage corpus, isolating answer quality given relevant evidence.

\paragraph{Inherent risk assessment (IRA)} This is a core audit-planning task: auditors must flag high-risk areas needing scrutiny without over-auditing low-risk ones. It is organized around inherent risk factors (IRFs), a standardized set of risk categories indicating where material misstatement or audit complexity may arise. For each \texttt{(account group, FSLI, IRF)} tuple, the pipeline assigns a risk level and produces an evidence-grounded justification from retrieved client documents.

\subsection{Systems Compared}
\label{sec:systems_compared}

We compare Litmus with four baseline configurations that automate final-output evaluation or rubric construction. Table~\ref{tab:comparison_protocol} summarizes the role and inputs of each system.

\begin{table}[t]
\centering
\setlength{\tabcolsep}{3pt}
\begin{tabular}{lcccc}
\toprule
System & Code & Labels & Scope & Mon. \\
\midrule
Litmus & yes & no & per-stage & yes \\
AutoMetrics & no & yes & final & no \\
DynRub-\textsc{Inst} & no & no & instance & no \\
DynRub-\textsc{DS} & no & no & dataset & no \\
DynRub-\textsc{FT} & no & feedback & final & no \\
\bottomrule
\end{tabular}
\caption{
Comparison protocol. Litmus is source-code grounded and produces a per-stage metric portfolio, while the baselines are final-output evaluation or rubric systems. Mon. = monitoring metrics; FT = fine-tuned.
}
\label{tab:comparison_protocol}
\end{table}

\begin{table*}[t]
\centering
\setlength{\tabcolsep}{4pt}
\resizebox{\textwidth}{!}{%
\begin{tabular}{lcccccccccc}
\toprule
& \multicolumn{6}{c}{Multi-axis (single-pass, temp.\ 0)} & \multicolumn{3}{c}{Single-vector validity (B3)} \\
\cmidrule(lr){2-7}\cmidrule(lr){8-10}
System & Cov.\ $\uparrow$ & Ground.\ $\uparrow$ & Redund.\ $\downarrow$ & Valid.\ $\tau\uparrow$ & Valid.\ $\rho\uparrow$ & Deg.\ ind.\ $\uparrow$ & $\rho$ [95\% CI] & $\tau$-b & $p$ \\
\midrule
\multicolumn{10}{l}{\emph{P1: Account grouping (financial), $n{=}112$}}\\
AutoMetrics & 0.30 & 0.41 & 0.14 & 0.30 & 0.34 & 0.54 & 0.45 [0.29, 0.59] & 0.35 & $<\!0.001$ \\
DynRub-\textsc{Inst} & 0.40 & 0.15 & \textbf{0.00} & 0.40 & 0.45 & 0.71 & 0.45 [0.29, 0.58] & 0.40 & $<\!0.001$ \\
DynRub-\textsc{DS} & 0.40 & 0.35 & \textbf{0.00} & 0.25 & 0.27 & 0.52 & 0.27 [0.08, 0.42] & 0.25 & $0.005$ \\
DynRub-\textsc{FT} & 0.40 & 0.35 & \textbf{0.00} & 0.41 & 0.47 & 0.68 & 0.47 [0.32, 0.60] & 0.41 & $<\!0.001$ \\
\textbf{Litmus (ours)} & \textbf{0.80} & \textbf{0.45} & 0.02 & \textbf{0.49} & \textbf{0.51} & \textbf{0.77} & \textbf{0.53 [0.37, 0.67]} & \textbf{0.49} & $<\!0.001$ \\
\midrule
\multicolumn{10}{l}{\emph{P2: Scientific QA (LitQA2), $n{=}58$}}\\
AutoMetrics & 0.67 & 0.19 & 0.14 & 0.68 & 0.69 & 0.92 & 0.47 [0.24, 0.63] & 0.42 & $<\!0.001$ \\
DynRub-\textsc{Inst} & 0.83 & 0.15 & \textbf{0.00} & $-0.06$ & $-0.06$ & \textbf{0.95} & $-0.06$ [$-0.12$, $-0.04$] & $-0.06$ & $1.00$ \\
DynRub-\textsc{DS} & 0.67 & \textbf{0.35} & \textbf{0.00} & $-0.12$ & $-0.12$ & \textbf{0.95} & $-0.12$ [$-0.23$, $-0.09$] & $-0.12$ & $1.00$ \\
DynRub-\textsc{FT} & \textbf{1.00} & \textbf{0.35} & \textbf{0.00} & 0.27 & 0.27 & \textbf{0.95} & 0.27 [$-0.07$, 0.71] & 0.27 & $0.040$ \\
\textbf{Litmus (ours)} & \textbf{1.00} & 0.32 & 0.01 & \textbf{0.71} & \textbf{0.72} & 0.94 & \textbf{0.72 [0.43, 0.93]} & \textbf{0.71} & $<\!0.001$ \\
\midrule
\multicolumn{10}{l}{\emph{P3: Inherent risk assessment (IRF), $n{=}112$}}\\
AutoMetrics & 0.83 & 0.16 & 0.07 & 0.25 & 0.31 & 0.71 & 0.36 [0.18, 0.51] & 0.27 & $<\!0.001$ \\
DynRub-\textsc{Inst} & 0.83 & 0.15 & \textbf{0.00} & 0.09 & 0.11 & 0.65 & 0.11 [$-0.08$, 0.28] & 0.09 & $0.245$ \\
DynRub-\textsc{DS} & 0.00 & 0.35 & \textbf{0.00} & 0.25 & 0.28 & 0.49 & 0.28 [0.10, 0.45] & 0.25 & $0.003$ \\
DynRub-\textsc{FT} & 0.83 & 0.35 & \textbf{0.00} & 0.21 & 0.25 & 0.82 & 0.25 [0.05, 0.42] & 0.21 & $0.008$ \\
\textbf{Litmus (ours)} & \textbf{0.83} & \textbf{0.42} & 0.05 & \textbf{0.27} & \textbf{0.32} & \textbf{0.84} & \textbf{0.39 [0.22, 0.52]} & \textbf{0.28} & $<\!0.001$ \\
\bottomrule
\end{tabular}%
}
\caption{Multi-axis comparison and single-vector validity (B3) across three
pipelines. Bold marks the best value per panel; arrows give the preferred
direction. Cov.~=~coverage, Ground.~=~grounding, Redund.~=~redundancy,
Valid.~$\tau$/$\rho$~=~validity vs.\ the per-row quality label, Deg.\ ind.~=~degradation
sensitivity (single-pass, temperature~0; the first three are LLM-assessed, see
\hyperref[sec:limitations]{Limitations}). B3 is the stitched single-vector score vs.\ the
same label, with bootstrap 95\% CIs and permutation $p$-values; its $\rho$ is a
single coefficient and differs from the per-metric Valid.\ $\rho$.}
\label{tab:main}
\end{table*}

Litmus emits per-stage portfolios (14 metrics for account grouping, 13 for scientific QA, 12 for inherent risk assessment) that include monitoring metrics, scope-matched LLM-as-judge metrics, and code-evaluated checks. The baselines (Section~\ref{sec:related}) instead score the final output and, unlike Litmus, produce no per-stage monitoring metrics; AutoMetrics is additionally trained on a separate client/train split, so its validity is out-of-distribution.

\subsection{Evaluation Axes}
\label{sec:evaluation_axes}

We score every system under a single protocol applied uniformly across all three domains, computing each axis identically for all systems, including ours, so no axis advantages a design by construction. We adopt two measures from the measurement-validity framework of \citet{ryan2026autometrics}: \emph{validity}, their criterion validity, measuring agreement between system scores and independent per-row quality labels; and \emph{degradation sensitivity}, their robustness sensitivity measure, testing whether a metric penalizes synthetically degraded outputs. We also report three portfolio-level \emph{design-quality} axes---\emph{coverage}, \emph{grounding}, and \emph{redundancy}---that characterize \emph{what} a metric-design system emits. Because final-output baselines were not built to optimize these axes, we treat them as descriptive diagnostics rather than a like-for-like contest. All LLM-based scoring (metric scores, the P1/P3 proxy labels, and the design-quality axes) uses temperature 0 with up to three retries and the same judge-model family. Section~\ref{sec:crossfamily} re-scores both the validity axis and the design-quality axes under two further families. Each axis is defined below; extended axis definitions and the five-level LLM-as-judge anchors are in Appendix~\ref{app:evaluation_axes}.

\paragraph{Validity} This is the \emph{criterion validity} of \citet{ryan2026autometrics}: the rank correlation (Kendall's $\tau$, Spearman's $\rho$) between a system's score on \emph{uncorrupted} (real, un-degraded) outputs and an independent \emph{per-row quality label}, one quality judgment per input--output row, used only as the external target. A \emph{scope-matched judge} is compared against the label for the stage it applies to (a retrieval judge against a retrieval-quality label, not a whole-pipeline one); for aggregate reporting we stitch a per-row score (Section~\ref{sec:results}).

\paragraph{Scoring judge and validity label} The \emph{scoring judge} applies a proposed rubric to an output and produces the score under evaluation; the \emph{validity label} is the separate target that score is compared against. On P2 the label is external LitQA2 gold \citep{skarlinski2024languageagentsachievesuperhuman}; on P1 and P3 it is an LLM-derived proxy scoped by the pipelines' routing labels and practitioner-confirmed component boundaries. Litmus uses none of these labels at design time; only the label-fitted AutoMetrics consumes them.

\paragraph{Design-quality and robustness axes} \emph{Coverage} is the fraction of a fixed, domain-specific list of reference failure concerns (ten for account grouping, six each for the others) addressed by at least one metric. \emph{Grounding} scores, on $[0,1]$, how concretely a metric refers to implementation artifacts (named modules, fields, rules, branches) rather than generic language. \emph{Redundancy} is the fraction of metric pairs judged to measure substantially the same behavior (lower is better); it is meaningful only for multi-metric portfolios, so the single-judge DynamicRubric variants are trivially non-redundant. \emph{Degradation sensitivity} is AutoMetrics' robustness check: whether a metric assigns lower scores to synthetically degraded outputs (10 rows over five severity levels per domain).

\section{Results}
\label{sec:results}

Because every Litmus metric is conditioned on the recovered component graph, we check that graph rather than assume it. On scientific QA, Litmus recovers a 30-node graph that is 65\% edge-verified against the tree-sitter call graph, with no phantom modules and an adversarial-critic confidence of 0.78 (Table~\ref{tab:archfidelity-pqa}, Appendix~\ref{app:arch}). 

On this footing, under a zero-label design-time constraint Litmus has the most favorable validity ranking: first on all three pipelines ($\rho=0.51, 0.72, 0.32$), ahead of every baseline including the label-fitted AutoMetrics, clearest on scientific QA and within overlapping intervals of the strongest baseline on P1 and P3 (Table~\ref{tab:main}). The stitched single-vector statistic keeps this order, with Litmus first and individually significant throughout ($p<0.001$). Uncapped coverage is top or tied on every pipeline, but that lead survives a size-matched comparison only on account grouping (Section~\ref{sec:sizecontrolled}). Litmus uniquely produces per-stage and operational metrics rather than scoring the final output alone; its \emph{redundancy} is near zero throughout, well below AutoMetrics; and on \emph{grounding} it leads on P1 and P3, trailing only slightly on P2. Litmus's judges are also among the most sensitive to corrupted outputs---top or tied on degradation on P1 and P3, and on par with the rubric systems on P2 (0.94 vs.\ 0.95)---so it is the only portfolio that is strong on both validity and degradation sensitivity across all three pipelines.

\section{Robustness}
\label{sec:robustness}

\subsection{Cross-Family Judge Control}
\label{sec:crossfamily}
To rule out a system being graded by a relative of itself, we re-score the held-out P2 rows---the one pipeline with external gold (LitQA2, $n{=}58$)---with OpenAI GPT-5.5 and GPT-4o, holding metrics, outputs, and gold labels fixed. The control covers every LLM-judge system; AutoMetrics is code-based and judge-invariant. Litmus stays first, its Spearman correlation to LitQA2 gold rising from $0.72$ under Claude to $0.86$ under either OpenAI judge (Table~\ref{tab:crossfamily}). The DynamicRubric baselines swing sharply with the assessor---the two rubric-only variants move from slightly negative under Claude to $+0.56$--$0.75$ under OpenAI, and the fine-tuned variant from $0.27$ to $0.48$--$0.56$---so their ordering is largely an artifact of the judge. We do \emph{not} claim judge-invariance on the proxy-labeled P1/P3.

The same worry applies to coverage, grounding, and redundancy. Re-scoring every portfolio under GPT-4o and GPT-5.5 leaves the ordering intact: Litmus stays ahead of AutoMetrics on grounding and redundancy in all nine pipeline--judge cells, and is first or tied on coverage in eight of nine (Table~\ref{tab:crossfamilyaxes}, Appendix~\ref{app:axescontrol}). Absolute scores shift with the assessor; the ranking does not.

\begin{table}[t]
\centering
\small
\setlength{\tabcolsep}{5pt}
\begin{tabular}{lccc}
\toprule
System & Claude & GPT-5.5 & GPT-4o \\
\midrule
AutoMetrics & 0.69 & 0.69 & 0.69 \\
DynRub-\textsc{Inst} & $-0.06$ & 0.75 & 0.56 \\
DynRub-\textsc{DS} & $-0.12$ & 0.67 & 0.69 \\
DynRub-\textsc{FT} & 0.27 & 0.48 & 0.56 \\
\textbf{Litmus} & \textbf{0.72} & \textbf{0.86} & \textbf{0.86} \\
\bottomrule
\end{tabular}
\caption{Cross-family judge control on scientific QA (P2, $n{=}58$): validity (Spearman $\rho$ vs.\ LitQA2 gold) under three judge families. Litmus stays first under two disjoint OpenAI judges while the rubric baselines swing with the judge; AutoMetrics is code-based and judge-invariant. Per-row Claude--OpenAI agreement: $\rho=0.64$, $\tau=0.63$, $\kappa\approx0.5$.}
\label{tab:crossfamily}
\end{table}

\subsection{Size-Controlled Coverage}
\label{sec:sizecontrolled}
Coverage grows with portfolio size, and Litmus emits more metrics than the baselines, so part of its coverage lead could be mechanical. We cap Litmus at AutoMetrics' eight metrics and recompute coverage two ways: a critic-verdict-ranked top-$k$ (pass $>$ flag $>$ unscored $>$ reject) and the mean over six random size-eight subsets (Table~\ref{tab:sizecontrolled}). On P1 the gap remains: $0.70$ ranked / $0.65\pm0.08$ random versus AutoMetrics' $0.30$. On P2 the full-size $1.00$ falls to a tie at $0.67$, and P3 is a tie at every size ($0.83$). Only P1 keeps a coverage lead at matched size.

\subsection{Adversarial-Validity Gate Ablation}
\label{sec:gateablation}
On the P2 draft the gate rejects $11$ of $77$ metrics and flags $10$. Dropping the rejects leaves coverage unchanged ($1.00$), cuts redundancy by $35\%$ ($0.024\rightarrow0.016$), and raises grounding ($0.377\rightarrow0.393$; Table~\ref{tab:gate}). The rejects are pipeline-mismatched (retrieval metrics on a PDF reader that does no query-time retrieval); the one scoreable metric is untouched ($\rho=0.72$).

\subsection{Reconstruction Fidelity vs.\ Metric Validity}
\label{sec:fidelityvalidity}
We pair each scoreable metric's validity with the host subsystem's reconstruction fidelity (Table~\ref{tab:fidelityvalidity}). Aggregate fidelity does not predict validity: P1 has higher edge-verification than P2 ($83\%$ vs.\ $65\%$) yet lower headline validity, and the pooled correlation is near zero (Spearman $-0.06$, $n{=}7$; component-level $-0.50$, $n{=}3$---neither powered). The gate still drops metrics whose reconstructed responsibility cannot support them (Section~\ref{sec:gateablation}). The lowest-confidence P2 node (\texttt{settings-config}, $0.65$) is exactly where the critic flagged the metric scoped there. With only $1$--$4$ scoreable metrics per pipeline a powered slope is not estimable.

These checks support a branch-diversity hypothesis, not an established predictor: elicitation helps most when a pipeline has multiple decision paths whose intended behavior is not recoverable from code---abstention semantics, fallback behavior, and a recall-versus-precision preference are the recurring cases on P2---and least when one dominant branch carries the signal (P1 and P3), where the validity intervals overlap the strongest baseline.

\section{Conclusion and Future Work}
\label{sec:conclusion}

We reframed automatic metric design as goal elicitation by interrogation: rather than selecting metrics first, Litmus interrogates a pipeline's source code and its practitioner to make the evaluation target explicit, then admits, rejects, or re-scopes each candidate against the resolved constraints. Across three domains this zero-label procedure yields near-zero redundancy and ranks first on label validity, clearest on scientific QA, with a coverage advantage on account grouping (Section~\ref{sec:sizecontrolled}); both that ranking on scientific QA and the design-quality picture on all three hold under a different judge family (Section~\ref{sec:robustness}). Key next steps are a controlled human study of portfolios generated with and without practitioner interrogation on the same pipeline, then independently rated and repaired, with rating deltas and edit distances as the outcomes; a full regeneration ablation of the four un-ablated stages (\hyperref[sec:limitations]{Limitations}); and broader pipelines.

\section*{Limitations}
\label{sec:limitations}

First, scope. Our \textbf{empirical evaluation} spans three real, code-defined pipelines from different domains---account grouping, scientific QA, and inherent risk assessment---primarily scored by a single judge-model family. We ran a cross-family judge control on the one pipeline with external gold labels (P2), where Litmus's validity ranking holds under two disjoint OpenAI judges (Section~\ref{sec:robustness}); on the proxy-labeled pipelines (P1, P3), whose validity labels are LLM-derived, judge-system alignment is not ruled out, so those results should still be read as evidence within one assessor family. Litmus itself is pipeline-agnostic, with no assumptions specific to any one domain, and broader multi-judge evaluation remains future work. Second, the systems are not strictly commensurable: Litmus is a zero-label metric-design system, whereas AutoMetrics is label-fitted and DynamicRubric is rubric-generating, and both score only the final output. AutoMetrics is fit on a separate client dataset and evaluated on the held-out validity rows, so its reported figure is already an out-of-distribution estimate rather than an optimistic in-distribution one. We treat this role difference as intentional rather than a confound, but it means absolute numbers should be read as characterizing what each system produces; the defensible claim is the ranking under a shared zero-label, design-time constraint. Third, three of our axes (coverage, grounding, and redundancy) are themselves LLM-assessed, and primarily by the model family that powers Litmus, risking judge--system alignment bias \citep{zheng2023judging,panickssery2024llm}; we mitigate this with a broad, fixed set of reference concerns per pipeline, and re-scoring every portfolio on all three pipelines under two further judge families leaves the ordering against AutoMetrics intact on grounding and redundancy in every pipeline--judge cell and on coverage in eight of nine (Section~\ref{sec:crossfamily}, Appendix~\ref{app:axescontrol}). Absolute scores still inherit the assessor's biases---one judge raises a baseline's account-grouping coverage from $0.30$ to $0.80$---so the ranking is again the more defensible claim. Fourth, coverage is monotone in portfolio size---Litmus emits a substantially larger portfolio than the baselines (14 metrics versus 1--8 on account grouping: AutoMetrics emits 8 and DynamicRubric a single holistic rubric score)---and the reference concern lists were authored by us. A size-controlled analysis (Section~\ref{sec:sizecontrolled}) shows the advantage survives capping only on P1. Reconstruction fidelity does not predict downstream validity in this sample (Section~\ref{sec:fidelityvalidity}), and only one of the five design stages is ablated: Section~\ref{sec:gateablation} isolates the adversarial-validity gate, while reconstruction, graph verification, clarification, and downstream metric design remain un-ablated, since a toggle-based ablation would require regenerating both arms of a non-deterministic synthesis step. Finally, the clarification channel's end-to-end benefit is shown here only at the specification level: we do not show that an answer changes which metrics enter the portfolio, or that validity improves as a result. Operational use does not establish improvement over prior manual or heuristic metric design; the controlled human study that would establish it, and a longitudinal evaluation of whether re-running Litmus on changed code keeps a portfolio current, both remain future work.

\section*{Ethical Considerations}
\label{sec:ethics}

Litmus analyzes source code and produces evaluation specifications; it does not make end-user-facing decisions, and the designed metrics are intended to inform practitioners rather than to act autonomously. Two of our three evaluation pipelines (account grouping and inherent risk assessment) are real audit workflows operating over proprietary client financial data; we do not release that data, identifiers are sanitized, and only aggregate measurements are reported. Because these are high-stakes audit and financial settings, automatically designed metrics should support---never replace---professional judgment and existing review controls. Designed metrics, particularly LLM-as-judge metrics, can encode the biases of the underlying model and should be reviewed and calibrated by practitioners before deployment. Finally, because metrics influence which system behaviours are surfaced and acted upon, over-trust in any automatically designed metric---including ours---risks masking failure modes it does not cover; the coverage axis is intended to make such gaps visible rather than to certify completeness.

\bibliography{custom}

@article{mulla2023automatic,
  title={Automatic question generation: a review of methodologies, datasets, evaluation metrics, and applications},
  author={Mulla, Nikahat and Gharpure, Prachi},
  journal={Progress in Artificial Intelligence},
  volume={12},
  number={1},
  pages={1--32},
  year={2023},
  publisher={Springer},
  doi={10.1007/s13748-023-00295-9},
  url={https://doi.org/10.1007/s13748-023-00295-9}
}

@inproceedings{ijcai2024p889,
  title     = {A Survey on Neural Question Generation: Methods, Applications, and Prospects},
  author    = {Guo, Shasha and Liao, Lizi and Li, Cuiping and Chua, Tat-Seng},
  booktitle = {Proceedings of the Thirty-Third International Joint Conference on
               Artificial Intelligence, {IJCAI-24}},
  publisher = {International Joint Conferences on Artificial Intelligence Organization},
  editor    = {Kate Larson},
  pages     = {8038--8047},
  year      = {2024},
  month     = {8},
  note      = {Survey Track},
  doi       = {10.24963/ijcai.2024/889},
  url       = {https://doi.org/10.24963/ijcai.2024/889},
}

@Inbook{Flor2025,
  author="Flor, Michael",
  title="Question Generation with Large Language Models and Generative AI",
  bookTitle="Automatic Question Generation",
  year="2025",
  publisher="Springer Nature Switzerland",
  address="Cham",
  pages="137--147",
  isbn="978-3-031-92072-1",
  doi="10.1007/978-3-031-92072-1_10",
  url="https://doi.org/10.1007/978-3-031-92072-1_10"
}

@inproceedings{desmond2025evalassist,
  title={Evalassist: Llm-as-a-judge simplified},
  author={Desmond, Michael and Ashktorab, Zahra and Geyer, Werner and Daly, Elizabeth M and Cooper, Martin Santillan and Pan, Qian and Nair, Rahul and Wagner, Nico and Pedapati, Tejaswini},
  booktitle={Proceedings of the AAAI Conference on Artificial Intelligence},
  volume={39},
  number={28},
  pages={29637--29639},
  year={2025},
  doi={10.1609/aaai.v39i28.35351},
  url={https://doi.org/10.1609/aaai.v39i28.35351}
}

@article{zheng2023judging,
  title={Judging llm-as-a-judge with mt-bench and chatbot arena},
  author={Zheng, Lianmin and Chiang, Wei-Lin and Sheng, Ying and Zhuang, Siyuan and Wu, Zhanghao and Zhuang, Yonghao and Lin, Zi and Li, Zhuohan and Li, Dacheng and Xing, Eric and others},
  journal={Advances in neural information processing systems},
  volume={36},
  pages={46595--46623},
  year={2023},
  doi={10.52202/075280-2020},
  url={https://doi.org/10.52202/075280-2020}
}

@article{panickssery2024llm,
  title={Llm evaluators recognize and favor their own generations},
  author={Panickssery, Arjun and Bowman, Samuel R and Feng, Shi},
  journal={Advances in Neural Information Processing Systems},
  volume={37},
  pages={68772--68802},
  year={2024},
  doi={10.52202/079017-2197},
  url={https://doi.org/10.52202/079017-2197}
}

@inproceedings{NIPS2015_86df7dcf,
 author = {Sculley, D. and Holt, Gary and Golovin, Daniel and Davydov, Eugene and Phillips, Todd and Ebner, Dietmar and Chaudhary, Vinay and Young, Michael and Crespo, Jean-Fran\c{c}ois and Dennison, Dan},
 booktitle = {Advances in Neural Information Processing Systems},
 editor = {C. Cortes and N. Lawrence and D. Lee and M. Sugiyama and R. Garnett},
 pages = {},
 publisher = {Curran Associates, Inc.},
 title = {Hidden Technical Debt in Machine Learning Systems},
 url = {https://proceedings.neurips.cc/paper_files/paper/2015/file/86df7dcfd896fcaf2674f757a2463eba-Paper.pdf},
 volume = {28},
 year = {2015}
}

@INPROCEEDINGS{8258038,
  author={Breck, Eric and Cai, Shanqing and Nielsen, Eric and Salib, Michael and Sculley, D.},
  booktitle={2017 IEEE International Conference on Big Data (Big Data)}, 
  title={The ML test score: A rubric for ML production readiness and technical debt reduction}, 
  year={2017},
  volume={},
  number={},
  pages={1123-1132},
  doi={10.1109/BigData.2017.8258038},
  url={https://doi.org/10.1109/BigData.2017.8258038}
}

@INPROCEEDINGS{8804457,
  author={Amershi, Saleema and Begel, Andrew and Bird, Christian and DeLine, Robert and Gall, Harald and Kamar, Ece and Nagappan, Nachiappan and Nushi, Besmira and Zimmermann, Thomas},
  booktitle={2019 IEEE/ACM 41st International Conference on Software Engineering: Software Engineering in Practice (ICSE-SEIP)}, 
  title={Software Engineering for Machine Learning: A Case Study}, 
  year={2019},
  volume={},
  number={},
  pages={291-300},
  doi={10.1109/ICSE-SEIP.2019.00042},
  url={https://doi.org/10.1109/ICSE-SEIP.2019.00042}
}

@inproceedings{10.1145/3287560.3287596,
  author = {Mitchell, Margaret and Wu, Simone and Zaldivar, Andrew and Barnes, Parker and Vasserman, Lucy and Hutchinson, Ben and Spitzer, Elena and Raji, Inioluwa Deborah and Gebru, Timnit},
  title = {Model Cards for Model Reporting},
  year = {2019},
  isbn = {9781450361255},
  publisher = {Association for Computing Machinery},
  address = {New York, NY, USA},
  url = {https://doi.org/10.1145/3287560.3287596},
  doi = {10.1145/3287560.3287596},
  booktitle = {Proceedings of the Conference on Fairness, Accountability, and Transparency},
  pages = {220–229},
  numpages = {10},
  location = {Atlanta, GA, USA},
  series = {FAT* '19}
}

@article{10.1145/3458723,
  author = {Gebru, Timnit and Morgenstern, Jamie and Vecchione, Briana and Vaughan, Jennifer Wortman and Wallach, Hanna and III, Hal Daum{\'e} and Crawford, Kate},
  title = {Datasheets for datasets},
  year = {2021},
  issue_date = {December 2021},
  publisher = {Association for Computing Machinery},
  address = {New York, NY, USA},
  volume = {64},
  number = {12},
  issn = {0001-0782},
  url = {https://doi.org/10.1145/3458723},
  doi = {10.1145/3458723},
  journal = {Commun. ACM},
  month = nov,
  pages = {86–92},
  numpages = {7}
}

@inproceedings{10.1145/3351095.3372873,
  author = {Raji, Inioluwa Deborah and Smart, Andrew and White, Rebecca N. and Mitchell, Margaret and Gebru, Timnit and Hutchinson, Ben and Smith-Loud, Jamila and Theron, Daniel and Barnes, Parker},
  title = {Closing the AI accountability gap: defining an end-to-end framework for internal algorithmic auditing},
  year = {2020},
  isbn = {9781450369367},
  publisher = {Association for Computing Machinery},
  address = {New York, NY, USA},
  url = {https://doi.org/10.1145/3351095.3372873},
  doi = {10.1145/3351095.3372873},
  booktitle = {Proceedings of the 2020 Conference on Fairness, Accountability, and Transparency},
  pages = {33–44},
  numpages = {12},
  location = {Barcelona, Spain},
  series = {FAT* '20}
}

@misc{wang2026generatingrefiningdynamicevaluation,
  title={Generating and Refining Dynamic Evaluation Rubrics for LLM-as-a-Judge}, 
  author={Zijie Wang and Eduardo Blanco},
  year={2026},
  eprint={2605.30568},
  archivePrefix={arXiv},
  primaryClass={cs.CL},
  doi={10.48550/arXiv.2605.30568},
  url={https://arxiv.org/abs/2605.30568}, 
}

@misc{skarlinski2024languageagentsachievesuperhuman,
  title={Language agents achieve superhuman synthesis of scientific knowledge}, 
  author={Michael D. Skarlinski and Sam Cox and Jon M. Laurent and James D. Braza and Michaela Hinks and Michael J. Hammerling and Manvitha Ponnapati and Samuel G. Rodriques and Andrew D. White},
  year={2024},
  eprint={2409.13740},
  archivePrefix={arXiv},
  primaryClass={cs.CL},
  url={https://arxiv.org/abs/2409.13740},
  doi={10.48550/arXiv.2409.13740}
}

@inproceedings{yao2023react,
  title={ReAct: Synergizing Reasoning and Acting in Language Models},
  author={Shunyu Yao and Jeffrey Zhao and Dian Yu and Nan Du and Izhak Shafran and Karthik R Narasimhan and Yuan Cao},
  booktitle={The Eleventh International Conference on Learning Representations },
  year={2023},
  url={https://openreview.net/forum?id=WE_vluYUL-X}
}

@inproceedings{ryan2026autometrics,
  title={AutoMetrics: Approximate Human Judgments with Automatically Generated Evaluators},
  author={Michael J Ryan and Yanzhe Zhang and Amol Salunkhe and Yi Chu and Di Xu and Diyi Yang},
  booktitle={The Fourteenth International Conference on Learning Representations},
  year={2026},
  url={https://openreview.net/forum?id=ymJuBifPUy}
}

@inproceedings{qin2024toolllm,
  title={Tool{LLM}: Facilitating Large Language Models to Master 16000+ Real-world {API}s},
  author={Yujia Qin and Shihao Liang and Yining Ye and Kunlun Zhu and Lan Yan and Yaxi Lu and Yankai Lin and Xin Cong and Xiangru Tang and Bill Qian and Sihan Zhao and Lauren Hong and Runchu Tian and Ruobing Xie and Jie Zhou and Mark Gerstein and dahai li and Zhiyuan Liu and Maosong Sun},
  booktitle={The Twelfth International Conference on Learning Representations},
  year={2024},
  url={https://openreview.net/forum?id=dHng2O0Jjr}
}

@inproceedings{liu2024agentbench,
  title={AgentBench: Evaluating {LLM}s as Agents},
  author={Xiao Liu and Hao Yu and Hanchen Zhang and Yifan Xu and Xuanyu Lei and Hanyu Lai and Yu Gu and Hangliang Ding and Kaiwen Men and Kejuan Yang and Shudan Zhang and Xiang Deng and Aohan Zeng and Zhengxiao Du and Chenhui Zhang and Sheng Shen and Tianjun Zhang and Yu Su and Huan Sun and Minlie Huang and Yuxiao Dong and Jie Tang},
  booktitle={The Twelfth International Conference on Learning Representations},
  year={2024},
  url={https://openreview.net/forum?id=zAdUB0aCTQ}
}

@inproceedings{zhou2024webarena,
  title={WebArena: A Realistic Web Environment for Building Autonomous Agents},
  author={Shuyan Zhou and Frank F. Xu and Hao Zhu and Xuhui Zhou and Robert Lo and Abishek Sridhar and Xianyi Cheng and Tianyue Ou and Yonatan Bisk and Daniel Fried and Uri Alon and Graham Neubig},
  booktitle={The Twelfth International Conference on Learning Representations},
  year={2024},
  url={https://openreview.net/forum?id=oKn9c6ytLx}
}

@inproceedings{jimenez2024swebench,
  title={{SWE}-bench: Can Language Models Resolve Real-world Github Issues?},
  author={Carlos E Jimenez and John Yang and Alexander Wettig and Shunyu Yao and Kexin Pei and Ofir Press and Karthik R Narasimhan},
  booktitle={The Twelfth International Conference on Learning Representations},
  year={2024},
  url={https://openreview.net/forum?id=VTF8yNQM66}
}

@inproceedings{fu-etal-2024-gptscore,
    title = "{GPTS}core: Evaluate as You Desire",
    author = "Fu, Jinlan  and
      Ng, See-Kiong  and
      Jiang, Zhengbao  and
      Liu, Pengfei",
    editor = "Duh, Kevin  and
      Gomez, Helena  and
      Bethard, Steven",
    booktitle = "Proceedings of the 2024 Conference of the North American Chapter of the Association for Computational Linguistics: Human Language Technologies (Volume 1: Long Papers)",
    month = jun,
    year = "2024",
    address = "Mexico City, Mexico",
    publisher = "Association for Computational Linguistics",
    url = "https://aclanthology.org/2024.naacl-long.365/",
    doi = "10.18653/v1/2024.naacl-long.365",
    pages = "6556--6576"
}

@inproceedings{schmidtova-etal-2024-automatic-metrics,
    title = "Automatic Metrics in Natural Language Generation: A survey of Current Evaluation Practices",
    author = "Schmidtova, Patricia  and
      Mahamood, Saad  and
      Balloccu, Simone  and
      Dusek, Ondrej  and
      Gatt, Albert  and
      Gkatzia, Dimitra  and
      Howcroft, David M.  and
      Platek, Ondrej  and
      Sivaprasad, Adarsa",
    editor = "Mahamood, Saad  and
      Minh, Nguyen Le  and
      Ippolito, Daphne",
    booktitle = "Proceedings of the 17th International Natural Language Generation Conference",
    month = sep,
    year = "2024",
    address = "Tokyo, Japan",
    publisher = "Association for Computational Linguistics",
    url = "https://aclanthology.org/2024.inlg-main.44/",
    doi = "10.18653/v1/2024.inlg-main.44",
    pages = "557--583"
}

@inproceedings{rao-daume-iii-2018-learning,
    title = "Learning to Ask Good Questions: Ranking Clarification Questions using Neural Expected Value of Perfect Information",
    author = "Rao, Sudha  and
      Daum{\'e} III, Hal",
    editor = "Gurevych, Iryna  and
      Miyao, Yusuke",
    booktitle = "Proceedings of the 56th Annual Meeting of the Association for Computational Linguistics (Volume 1: Long Papers)",
    month = jul,
    year = "2018",
    address = "Melbourne, Australia",
    publisher = "Association for Computational Linguistics",
    url = "https://aclanthology.org/P18-1255/",
    doi = "10.18653/v1/P18-1255",
    pages = "2737--2746"
}

@inproceedings{rao-daume-iii-2019-answer,
    title = "{A}nswer-based {A}dversarial {T}raining for {G}enerating {C}larification {Q}uestions",
    author = "Rao, Sudha  and
      Daum{\'e} III, Hal",
    editor = "Burstein, Jill  and
      Doran, Christy  and
      Solorio, Thamar",
    booktitle = "Proceedings of the 2019 Conference of the North {A}merican Chapter of the Association for Computational Linguistics: Human Language Technologies, Volume 1 (Long and Short Papers)",
    month = jun,
    year = "2019",
    address = "Minneapolis, Minnesota",
    publisher = "Association for Computational Linguistics",
    url = "https://aclanthology.org/N19-1013/",
    doi = "10.18653/v1/N19-1013",
    pages = "143--155"
}

@inproceedings{majumder-etal-2021-ask,
    title = "Ask what{'}s missing and what{'}s useful: Improving Clarification Question Generation using Global Knowledge",
    author = "Majumder, Bodhisattwa Prasad  and
      Rao, Sudha  and
      Galley, Michel  and
      McAuley, Julian",
    editor = "Toutanova, Kristina  and
      Rumshisky, Anna  and
      Zettlemoyer, Luke  and
      Hakkani-Tur, Dilek  and
      Beltagy, Iz  and
      Bethard, Steven  and
      Cotterell, Ryan  and
      Chakraborty, Tanmoy  and
      Zhou, Yichao",
    booktitle = "Proceedings of the 2021 Conference of the North American Chapter of the Association for Computational Linguistics: Human Language Technologies",
    month = jun,
    year = "2021",
    address = "Online",
    publisher = "Association for Computational Linguistics",
    url = "https://aclanthology.org/2021.naacl-main.340/",
    doi = "10.18653/v1/2021.naacl-main.340",
    pages = "4300--4312"
}

@inproceedings{liu-etal-2023-g,
    title = "{G}-Eval: {NLG} Evaluation using Gpt-4 with Better Human Alignment",
    author = "Liu, Yang  and
      Iter, Dan  and
      Xu, Yichong  and
      Wang, Shuohang  and
      Xu, Ruochen  and
      Zhu, Chenguang",
    editor = "Bouamor, Houda  and
      Pino, Juan  and
      Bali, Kalika",
    booktitle = "Proceedings of the 2023 Conference on Empirical Methods in Natural Language Processing",
    month = dec,
    year = "2023",
    address = "Singapore",
    publisher = "Association for Computational Linguistics",
    url = "https://aclanthology.org/2023.emnlp-main.153/",
    doi = "10.18653/v1/2023.emnlp-main.153",
    pages = "2511--2522"
}

@inproceedings{papineni-etal-2002-bleu,
    title = "{B}leu: a Method for Automatic Evaluation of Machine Translation",
    author = "Papineni, Kishore  and
      Roukos, Salim  and
      Ward, Todd  and
      Zhu, Wei-Jing",
    editor = "Isabelle, Pierre  and
      Charniak, Eugene  and
      Lin, Dekang",
    booktitle = "Proceedings of the 40th Annual Meeting of the Association for Computational Linguistics",
    month = jul,
    year = "2002",
    address = "Philadelphia, Pennsylvania, USA",
    publisher = "Association for Computational Linguistics",
    url = "https://aclanthology.org/P02-1040/",
    doi = "10.3115/1073083.1073135",
    pages = "311--318"
}

@inproceedings{lin-2004-rouge,
    title = "{ROUGE}: A Package for Automatic Evaluation of Summaries",
    author = "Lin, Chin-Yew",
    booktitle = "Text Summarization Branches Out",
    month = jul,
    year = "2004",
    address = "Barcelona, Spain",
    publisher = "Association for Computational Linguistics",
    url = "https://aclanthology.org/W04-1013/",
    pages = "74--81"
}

@article{reiter-2018-structured,
    title = "A Structured Review of the Validity of {BLEU}",
    author = "Reiter, Ehud",
    journal = "Computational Linguistics",
    volume = "44",
    number = "3",
    month = sep,
    year = "2018",
    address = "Cambridge, MA",
    publisher = "MIT Press",
    url = "https://aclanthology.org/J18-3002/",
    doi = "10.1162/coli_a_00322",
    pages = "393--401"
}

@inproceedings{zhou-etal-2022-deconstructing,
    title = "Deconstructing {NLG} Evaluation: Evaluation Practices, Assumptions, and Their Implications",
    author = "Zhou, Kaitlyn  and
      Blodgett, Su Lin  and
      Trischler, Adam  and
      Daum{\'e} III, Hal  and
      Suleman, Kaheer  and
      Olteanu, Alexandra",
    editor = "Carpuat, Marine  and
      de Marneffe, Marie-Catherine  and
      Meza Ruiz, Ivan Vladimir",
    booktitle = "Proceedings of the 2022 Conference of the North American Chapter of the Association for Computational Linguistics: Human Language Technologies",
    month = jul,
    year = "2022",
    address = "Seattle, United States",
    publisher = "Association for Computational Linguistics",
    url = "https://aclanthology.org/2022.naacl-main.24/",
    doi = "10.18653/v1/2022.naacl-main.24",
    pages = "314--324"
}

@inproceedings{jia-etal-2026-autorubric,
    title = "{A}uto{R}ubric: Rubric-Based Generative Rewards for Faithful Multimodal Reasoning",
    author = "Jia, Mengzhao  and
      Zhang, Zhihan  and
      Cases, Ignacio  and
      Liu, Zheyuan  and
      Jiang, Meng  and
      Qi, Peng",
    editor = "Liakata, Maria  and
      Moreira, Viviane P.  and
      Zhang, Jiajun  and
      Jurgens, David",
    booktitle = "Findings of the {A}ssociation for {C}omputational {L}inguistics: {ACL} 2026",
    month = jul,
    year = "2026",
    address = "San Diego, California, United States",
    publisher = "Association for Computational Linguistics",
    url = "https://aclanthology.org/2026.findings-acl.1282/",
    doi = "10.18653/v1/2026.findings-acl.1282",
    pages = "25707--25724",
    ISBN = "979-8-89176-395-1"
}

\appendix

\section{Reference Failure-Concerns}
\label{app:concerns}
The coverage axis uses a fixed list of reference failure-concerns per pipeline. For account grouping (ten): (1) account assigned to a semantically wrong group; (2) group inconsistent with / contradicting the stated FSLI; (3) business process context ignored; (4) high confidence on an incorrect assignment (miscalibration); (5) missing or empty reasoning; (6) grouping not grounded in retrieved evidence (hallucination); (7) wrong granularity of disaggregation; (8) override rules for account types not respected; (9) operational failure (exceptions/retries/degraded service); (10) retrieval quality issues (low recall / irrelevant chunks). For scientific QA (six): (1) chosen option does not match the correct answer; (2) answer not grounded in the retrieved evidence (hallucinated); (3) answer contradicts the cited passage; (4) relevant evidence present but ignored / not used; (5) overconfident answer when the evidence is insufficient; (6) incomplete answer that omits key parts of the question. For inherent risk assessment (six): (1) wrong risk severity level assigned; (2) missing or wrong risk category; (3) unsupported / hallucinated risk not grounded in input; (4) missing or low-quality mitigation recommendation; (5) inconsistent or contradictory risk justification; (6) relevant context not retrieved / used.

\section{Scoring and Implementation Details}
\label{sec:scoring_details}
The LLM scoring protocol (temperature 0, up to three retries, and one judge family for metric scores, P1/P3 proxy labels, and design-quality axes) is stated in Section~\ref{sec:evaluation_axes}. The five-level anchors used by Litmus's LLM-as-judge metrics are in Appendix~\ref{app:evaluation_axes}; the designer prompt that embeds them is in Appendix~\ref{app:impl}. Judge-family alignment bias is discussed in \hyperref[sec:limitations]{Limitations} and controlled for in Section~\ref{sec:crossfamily}; Appendix~\ref{app:axescontrol} reports the design-quality axes under three judge families.

The coverage reference lists contain ten concerns for account grouping and six each for scientific QA and inherent risk assessment (enumerated in Appendix~\ref{app:concerns}). For degradation sensitivity, each domain corrupts 10 rows across five cumulative severity levels.

\subsection{Evaluation Axes}
\label{app:evaluation_axes}

We score every system under a single protocol applied uniformly across all three domains, computing each axis identically for all systems---including ours---so that no axis advantages a particular design by construction. The axes serve two distinct purposes. \emph{Validity} is our label-grounded comparison: it tests whether a system's scores agree with an external per-row quality signal, the standard notion of whether an automatically produced metric or judge is trustworthy. \emph{Coverage}, \emph{grounding}, and \emph{redundancy} are portfolio-level \emph{design-quality} axes that we introduce to characterize \emph{what} an automatic metric-design system emits; they are not objectives the final-output baselines were built to optimize, so we report them as descriptive, diagnostic context rather than as a like-for-like contest. \emph{Degradation sensitivity} is a complementary robustness check that we report alongside validity: it confirms a system reacts to corrupted outputs, which validity alone does not establish.

\subsubsection{Validity}

Validity measures the association between a system's score on uncorrupted outputs and an independent per-row quality label, reported as Kendall's $\tau$ and Spearman's $\rho$. It is the only axis grounded in a target external to every system and is therefore our primary head-to-head comparison. The label is used only here: no system except AutoMetrics (which is label-fitted) consumes it, and Litmus uses none at design time. For branch- or stage-scoped metrics, each judge is evaluated against the scope-matched label; for aggregate reporting we also compute a stitched per-row score, as described in Section~\ref{sec:results}.

\subsubsection{Coverage}

Coverage measures the fraction of a fixed, domain-specific list of reference failure concerns addressed by at least one metric: ten reference concerns for account grouping, and six each for scientific QA and inherent risk assessment. It asks whether a portfolio spans the relevant failure surface rather than producing a single generic score. Because the final-output baselines emit one judge or rubric, coverage characterizes a property of portfolio breadth they were not designed to provide, and we report it descriptively.

\subsubsection{Grounding}

Grounding measures how concretely a metric refers to implementation artifacts---named modules, fields, rules, branches, or source-code behavior---rather than using generic evaluation language, scored on a $[0,1]$ scale with higher values indicating stronger grounding. It reflects Litmus's source-code-grounded design goal; baselines that score only the final output do not have access to such artifacts, so this axis describes artifact specificity rather than a deficiency the baselines could have avoided.

\subsubsection{Redundancy}

Redundancy measures the fraction of metric pairs judged to measure substantially the same behavior, with lower being better. It penalizes portfolios that appear broad but repeatedly measure the same property. The axis is meaningful only for multi-metric portfolios: single-judge baselines (the DynamicRubric variants) are trivially non-redundant, which we mark explicitly so the axis is not misread as an advantage.

\subsubsection{Degradation sensitivity}

Degradation sensitivity measures how much a metric's score drops when outputs are synthetically corrupted at increasing severity. A trustworthy metric must react to degraded outputs; we report degradation sensitivity alongside validity because the two are \emph{complementary}---sensitivity confirms that a metric responds when outputs are broken, while validity confirms that it tracks genuine quality on real outputs.

\paragraph{LLM-as-judge scoring scale.}
Litmus's LLM-as-judge metrics use one five-level scale. Every score is accompanied by a chain-of-thought rationale: the judge must explain its reasoning, consider both strengths and weaknesses, and then assign a score. The anchors, taken from the designer prompt in Appendix~\ref{app:impl}, are:
\begin{itemize}
  \item 5 (Excellent): Fully meets all criteria with no issues.
  \item 4 (Good): Meets criteria with only minor issues.
  \item 3 (Acceptable): Partially meets criteria with some issues.
  \item 2 (Poor): Largely fails to meet criteria with major issues.
  \item 1 (Very Poor): Fails to meet criteria entirely.
\end{itemize}

\subsection{Cross-Family Control on the Design-Quality Axes}
\label{app:axescontrol}

Coverage, grounding, and redundancy are scored from metric names and descriptions alone, by the same notebook functions used for Table~\ref{tab:main}. We held every portfolio fixed and re-ran those functions under Claude Opus~4.6 (the family that also powers Litmus), GPT-4o, and GPT-5.5, one execution per pipeline so the three columns of a pipeline are comparable. Across the two runs the three families issued $1{,}503$ judge calls; every response parsed, with no empty or off-format replies.

The Claude column is that re-run, not a copy of Table~\ref{tab:main}. Redundancy reproduced exactly on all three pipelines; grounding to within $0.008$. Coverage moved by one flipped concern on account grouping ($0.80\rightarrow 0.90$) and on scientific QA ($1.00\rightarrow 0.83$), and was exact on risk assessment. Coverage is ten binary items on the first pipeline and six on the other two, so a single flip is the smallest possible disagreement; judges are called at temperature 0 but are not bit-deterministic. Mixing the published Claude numbers with a fresh OpenAI column would put two executions of a non-deterministic judge side by side, so we do not.

\begin{table*}[t]
\centering
\small
\setlength{\tabcolsep}{4.5pt}
\begin{tabular}{l ccc ccc ccc}
\toprule
& \multicolumn{3}{c}{P1 Account Grouping} & \multicolumn{3}{c}{P2 PaperQA2/LitQA2} & \multicolumn{3}{c}{P3 Risk Assessment} \\
\cmidrule(lr){2-4}\cmidrule(lr){5-7}\cmidrule(lr){8-10}
System & Claude & GPT-4o & GPT-5.5 & Claude & GPT-4o & GPT-5.5 & Claude & GPT-4o & GPT-5.5 \\
\midrule
\multicolumn{10}{l}{\emph{Coverage} $\uparrow$} \\
AutoMetrics            & 0.300 & 0.800 & 0.300 & 0.833 & 1.000 & 1.000 & 0.833 & 0.833 & 0.833 \\
DynRub-\textsc{Inst}   & 0.500 & 0.300 & 0.400 & 0.667 & 1.000 & 0.667 & 0.667 & 0.833 & 0.667 \\
DynRub-\textsc{DS}     & 0.400 & 0.300 & 0.300 & 0.667 & 0.833 & 0.333 & 0.000 & 0.500 & 0.500 \\
DynRub-\textsc{FT}     & 0.500 & 0.800 & 0.200 & 0.833 & 1.000 & 0.500 & 0.667 & 1.000 & 0.500 \\
\textbf{Litmus}        & \textbf{0.900} & \textbf{1.000} & \textbf{0.900} & \textbf{0.833} & \textbf{1.000} & \textbf{0.833} & \textbf{0.833} & \textbf{1.000} & \textbf{0.833} \\
\midrule
\multicolumn{10}{l}{\emph{Grounding} $\uparrow$} \\
AutoMetrics            & 0.394 & 0.375 & 0.337 & 0.175 & 0.212 & 0.138 & 0.163 & 0.037 & 0.069 \\
DynRub-\textsc{Inst}   & 0.300$^{\dagger}$ & 0.000$^{\dagger}$ & 0.250$^{\dagger}$ & 0.150$^{\dagger}$ & 0.000$^{\dagger}$ & 0.350$^{\dagger}$ & 0.150$^{\dagger}$ & 0.000$^{\dagger}$ & 0.300$^{\dagger}$ \\
DynRub-\textsc{DS}     & 0.350$^{\dagger}$ & 0.300$^{\dagger}$ & 0.350$^{\dagger}$ & 0.350$^{\dagger}$ & 0.200$^{\dagger}$ & 0.350$^{\dagger}$ & 0.350$^{\dagger}$ & 0.300$^{\dagger}$ & 0.350$^{\dagger}$ \\
DynRub-\textsc{FT}     & 0.350$^{\dagger}$ & 0.700$^{\dagger}$ & 0.450$^{\dagger}$ & 0.250$^{\dagger}$ & 0.300$^{\dagger}$ & 0.350$^{\dagger}$ & 0.250$^{\dagger}$ & 0.300$^{\dagger}$ & 0.450$^{\dagger}$ \\
\textbf{Litmus}        & \textbf{0.450} & \textbf{0.536} & \textbf{0.399} & \textbf{0.331} & \textbf{0.331} & \textbf{0.269} & \textbf{0.417} & \textbf{0.375} & \textbf{0.323} \\
\midrule
\multicolumn{10}{l}{\emph{Redundancy} $\downarrow$} \\
AutoMetrics            & 0.143 & 0.179 & 0.143 & 0.179 & 0.179 & 0.107 & 0.071 & 0.036 & 0.036 \\
DynRub-\textsc{Inst}   & 0.000$^{\ddagger}$ & 0.000$^{\ddagger}$ & 0.000$^{\ddagger}$ & 0.000$^{\ddagger}$ & 0.000$^{\ddagger}$ & 0.000$^{\ddagger}$ & 0.000$^{\ddagger}$ & 0.000$^{\ddagger}$ & 0.000$^{\ddagger}$ \\
DynRub-\textsc{DS}     & 0.000$^{\ddagger}$ & 0.000$^{\ddagger}$ & 0.000$^{\ddagger}$ & 0.000$^{\ddagger}$ & 0.000$^{\ddagger}$ & 0.000$^{\ddagger}$ & 0.000$^{\ddagger}$ & 0.000$^{\ddagger}$ & 0.000$^{\ddagger}$ \\
DynRub-\textsc{FT}     & 0.000$^{\ddagger}$ & 0.000$^{\ddagger}$ & 0.000$^{\ddagger}$ & 0.000$^{\ddagger}$ & 0.000$^{\ddagger}$ & 0.000$^{\ddagger}$ & 0.000$^{\ddagger}$ & 0.000$^{\ddagger}$ & 0.000$^{\ddagger}$ \\
\textbf{Litmus}        & \textbf{0.022} & \textbf{0.011} & \textbf{0.011} & \textbf{0.013} & \textbf{0.000} & \textbf{0.000} & \textbf{0.045} & \textbf{0.000} & \textbf{0.000} \\
\bottomrule
\end{tabular}
\caption{Cross-family judge control on the three LLM-assessed design axes, all three pipelines. Portfolios are held fixed and re-scored by Claude Opus~4.6, GPT-4o, and GPT-5.5 in a single execution per pipeline. Against AutoMetrics, the only other multi-metric portfolio, Litmus is better on grounding and on redundancy in all nine pipeline--judge cells, and first or tied on coverage in eight of nine; the exception is scientific QA under GPT-5.5. Absolute levels shift with the assessor; the ordering does not. $\dagger$: single-metric portfolio, so grounding is the mean of one judge call. $\ddagger$: redundancy is defined over unordered metric pairs, so a one-metric portfolio scores $0.000$ by construction. Litmus portfolio sizes: 14 / 13 / 12; AutoMetrics 8 and each DynamicRubric variant 1 on every pipeline.}
\label{tab:crossfamilyaxes}
\end{table*}

\subsection{Size Control, Gate Ablation, and Reconstruction Fidelity}
\label{app:robustness-tables}

Tables for the size-controlled coverage check, the adversarial-validity gate ablation, and the reconstruction-fidelity pairing (Sections~\ref{sec:sizecontrolled}--\ref{sec:fidelityvalidity}).

\begin{table}[t]
\centering
\small
\setlength{\tabcolsep}{4pt}
\begin{tabular}{lcccc}
\toprule
Pipeline & Full & @8 rank & @8 rand & Auto. \\
\midrule
P1 Acct.\ grouping & 0.80 & 0.70 & $0.65{\pm}.08$ & 0.30 \\
P2 Scientific QA & 1.00 & 0.67 & $0.58{\pm}.32$ & 0.67 \\
P3 Risk assessment & 0.83 & 0.83 & $0.83{\pm}.00$ & 0.83 \\
\bottomrule
\end{tabular}
\caption{Size-controlled coverage. Litmus full portfolio vs.\ capped to $k{=}8$ (AutoMetrics' size) by verdict-ranking and by random size-8 subsets (mean$\pm$sd), vs.\ AutoMetrics. The coverage advantage survives capping only on P1; P2 and P3 tie at matched size.}
\label{tab:sizecontrolled}
\end{table}

\begin{table}[t]
\centering
\small
\setlength{\tabcolsep}{6pt}
\begin{tabular}{lccc}
\toprule
Axis & Ungated (77) & Gated (66) & $\Delta$ \\
\midrule
Coverage $\uparrow$ & 1.000 & 1.000 & $+0.000$ \\
Redundancy $\downarrow$ & 0.024 & 0.016 & $-35\%$ \\
Grounding $\uparrow$ & 0.377 & 0.393 & $+0.017$ \\
\bottomrule
\end{tabular}
\caption{Adversarial-validity gate ablation on scientific QA (P2). Counts are over the whole-solution drafted portfolio across all components, not the 13-metric scoped portfolio in Table~\ref{tab:main}. Removing the $11$ rejected metrics (of $77$ drafted; $10$ flagged) cuts redundancy by $35\%$ and raises grounding at no coverage cost; the one scoreable metric is unchanged ($\rho=0.72$).}
\label{tab:gate}
\end{table}

\begin{table}[t]
\centering
\small
\setlength{\tabcolsep}{3pt}
\begin{tabular}{lcccc}
\toprule
Pipeline & Edge-ver.\% & Conf.\ & Min & Validity $\rho$ \\
\midrule
P1 Acct.\ grouping & 83 & 0.93 & 0.90 & 0.51 \\
P2 Scientific QA & 65 & 0.83 & 0.65 & 0.72 \\
P3 Risk assessment & 83 & 0.92 & 0.88 & 0.32 \\
\bottomrule
\end{tabular}
\caption{Reconstruction fidelity vs.\ downstream validity. Edge-verification \% and host-subsystem critic confidence (mean, min) against headline validity $\rho$. P2's $0.83$ is the five-node \texttt{core-qa-engine} mean; whole-graph confidence across all 30 nodes is $0.78$ (Section~\ref{sec:results}). Aggregate fidelity does not track validity (pooled $\rho=-0.06$, $n{=}7$).}
\label{tab:fidelityvalidity}
\end{table}

\section{Architecture Generation: Pipeline and Protocol}
\label{app:arch}

\begin{figure*}[t]
\centering
\includegraphics[width=\textwidth]{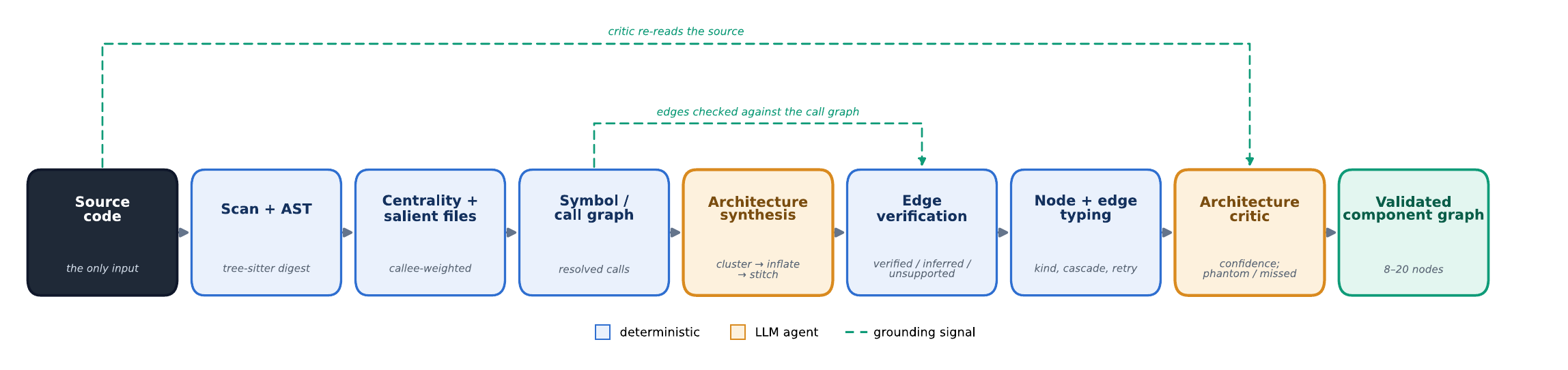}
\caption{Litmus's architecture-generation pipeline. Deterministic stages (blue) bracket the LLM stages (orange): static analysis and a symbol/call graph feed an LLM synthesizer, whose output is grounded by deterministic edge verification and an adversarial LLM critic before a validated component graph is emitted. Dashed green arrows are the grounding signals---synthesized edges are checked against the call graph, and the critic re-reads the source.}
\label{fig:litmus-arch-generation}
\end{figure*}

\paragraph{How Litmus reconstructs an architecture.}
Figure~\ref{fig:litmus-arch-generation} shows the generation pipeline. A deterministic front end scans the repository, extracts tree-sitter digests, scores files by callee-weighted centrality to pick salient ones, and builds a symbol/call graph. An LLM then synthesizes an 8--20 node component graph (clustering files into subsystems, inflating each into nodes, and stitching cross-subsystem edges), with every edge required to cite the calls that justify it. Two grounding steps follow: a deterministic pass marks each edge \emph{verified}, \emph{inferred}, or \emph{unsupported} by checking its citations against the call graph (dropping unsupported ones), and an adversarial LLM critic re-reads the source to score per-node confidence, prune phantoms, and flag misses. Only after these checks is the graph used for metric design, which is what lets us report fidelity as a measurement rather than an assumption.

\begin{figure*}[t!]
\centering
\includegraphics[width=\textwidth,height=0.82\textheight,keepaspectratio]{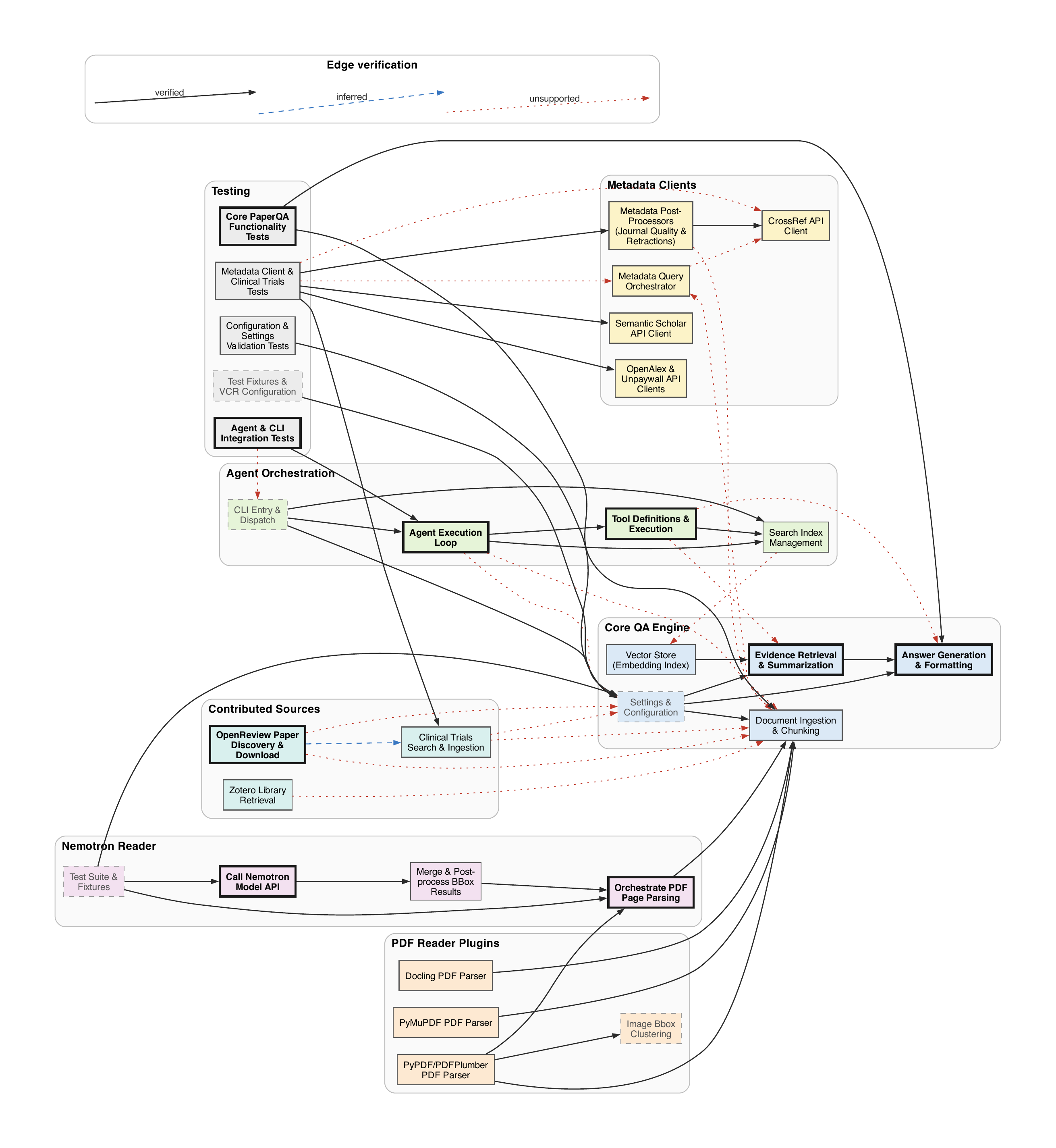}
\caption{Architecture graph Litmus reconstructs for the scientific-QA pipeline
(PaperQA2): 30 component nodes spanning seven subsystems---PDF reader plugins
(Docling, PyMuPDF, PyPDF/PDFPlumber), agent orchestration (CLI dispatch, execution
loop, tool definitions, search index), metadata clients (CrossRef, Semantic
Scholar, OpenAlex/Unpaywall, journal-quality and retraction post-processors), the
core QA engine (ingestion \& chunking, evidence retrieval \& summarization, answer
generation, vector store, settings), the Nemotron reader (page-parse
orchestration, Nemotron API call, bbox merge), contributed sources (Zotero,
OpenReview, clinical trials), and the test suite. Nodes and data-flow edges are
emitted by the synthesizer and grounded by deterministic edge verification and the
adversarial critic before any metric is designed.}
\label{fig:archpqa}
\end{figure*}

\paragraph{The graph recovered for account grouping.}
Running this pipeline on the audit pipeline yields the eight components in Table~\ref{tab:archgraph}: a trial-balance ingest point, an FSLI exact-matcher, three parallel mapping branches (account-type override, prioritized-RAG, disaggregated-FSLI), a shared Azure Cognitive Search client, an orchestrator that routes accounts and post-processes the result (resolving FSLI contradictions and tidying \texttt{account\_group}), and an output assembler. The three branches it finds are the same ones the validity analysis is scoped to, even though that scoping came from the routing labels and not from the graph.

\begin{table}[t]
\centering
\setlength{\tabcolsep}{3pt}
\resizebox{\columnwidth}{!}{%
\begin{tabular}{lllll}
\toprule
\# & Component & Role & Kind & Silent-fail \\
\midrule
1 & Trial-Balance Ingest        & entry         & compute       & low \\
2 & FSLI Exact Matcher          & supporting    & compute       & medium \\
3 & Account-Type Override Mapper & ai\_core      & prompt        & high \\
4 & Prioritized-RAG Mapper      & ai\_core      & compute       & high \\
5 & Disaggregated-FSLI Mapper   & ai\_core      & compute       & high \\
6 & Azure Cognitive Search      & supporting    & vector\_store & medium \\
7 & Orchestrator + Post-proc    & orchestrator  & compute       & high \\
8 & Output Assembler            & output        & compute       & medium \\
\bottomrule
\end{tabular}%
}
\caption{Components Litmus recovers for the account-grouping pipeline. Roles, kinds, and silent-failure risk are emitted fields of the synthesizer schema.}
\label{tab:archgraph}
\end{table}

\paragraph{How many edges are real.}
Each proposed edge cites caller$\to$callee symbol pairs that a deterministic pass checks against the tree-sitter call graph: \emph{verified} when the evidence resolves, \emph{inferred} when none is cited or found, \emph{unsupported} when the cited symbols are absent (and then discarded unless needed for connectivity). Here, of \textbf{49} edges, \textbf{32} are verified, \textbf{1} inferred, and \textbf{16} unsupported (\textbf{0} discarded, all retained for connectivity)---so \textbf{65\%} of the claimed data flow points to a function call we can actually locate.

\paragraph{What the critic caught.}
The critic scores its confidence in each node and proposes missing ones; nodes below $0.3$ flagged ``not found'' are removed automatically. On this pipeline it returned overall confidence \textbf{0.78} (lowest node confidence \textbf{0.65}), pruned \textbf{0} phantom nodes---no component's claimed files were absent---and flagged \textbf{3} missed, e.g.\ \emph{Types \& Data Models} (Table~\ref{tab:archfidelity-pqa}).

\paragraph{The evidence behind three components.}
Identifiers are sanitized; the data is proprietary (\hyperref[sec:ethics]{Ethical Considerations}). \emph{Override Mapper (3)} maps to the override-mapper module plus \texttt{override\_rules.json} (classes \textsc{asset, liability, equity, revenue, expense, non-operating, other-operating, non-controlling-interest}); verified edges include \texttt{orchestrator:route\_accounts $\to$ override\_mapper:classify\_batch} and \texttt{$\to$ llm\_client:complete}, which is why its judge is scoped to ``Account type override'' rows. \emph{RAG Mapper (4)} maps to the RAG module with verified edges \texttt{rag\_mapper:map $\to$ azure\_search\_client:search} and \texttt{$\to$ llm\_client:complete}, licensing a groundedness judge. \emph{Azure Search (6)}, detected as \texttt{vector\_store}, has three verified inbound edges (nodes 2, 4, 5), so one dependency-health monitor covers all three callers.

\paragraph{Self-review (adversarial validity) questions.} After drafting candidate metrics, Litmus interrogates each one along four axes ---pipeline fit, validity conditions, data assumptions, and direction of goodness ---and emits a verdict of \textsc{pass}, \textsc{flag}, or \textsc{reject}. On the scientific-QA pipeline the critic scored 57 of the 77 drafted metrics (20 were never scored): it passed 36, flagged 10, and rejected 11. Representative resolved questions:

\begin{itemize}
\item \textbf{Pipeline fit.} ``Does \emph{RAG Context Precision} measure the component it is attached to?'' \textsc{reject}: the component is a PDF reader/parser that extracts text; it performs no query-based retrieval, so a retrieval-quality metric cannot apply.
\item \textbf{Data assumptions.} ``Does \emph{Tool Selection Accuracy} have the ground truth it needs?'' \textsc{flag}: the computation requires a golden-tool mapping; it is unclear whether such labels exist in this codebase.
\item \textbf{Data assumptions.} ``Can \emph{RAG Retrieval Recall@K} on the retrieval stage be computed?''
\textsc{flag}: Recall@K needs known-relevant documents per query, and no labeled evaluation set was found.
\item \textbf{Validity conditions.} ``Is the \emph{Settings \& Configuration} tier really an LLM-generation stage?'' \textsc{flag}: it manages prompt templates and configuration rather than generating text, so the \{LLM\} classification is questionable.
\item \textbf{Pipeline fit.} ``Do production-monitoring metrics (\emph{Error Rate per Exception Class}, \emph{Retry Success Rate}, \emph{Dead-Letter Queue Rate}) apply to the test-suite component?'' \textsc{flag}: test suites are not production services, so exception/retry rates indicate test failures, not production health.
\item \textbf{Redundancy.} ``Is \emph{Exception Recovery Rate} distinct from \emph{Graceful Degradation Success Rate}?'' \textsc{flag}: both measure whether caught exceptions still yield usable responses; the distinction is thin.
\end{itemize}

\paragraph{Practitioner clarification questions.}
Litmus also drafts clarification questions for the practitioner, each emitted with an explicit \emph{metric-impact} statement of what its answer changes. The following are representative questions for the scientific-QA components (the direction-of-goodness and scoping decisions they resolve are noted in italics):

\begin{itemize}
\item \textbf{Evidence Retrieval \& Summarization:} ``Is this stage tuned for broad recall or precise grounding when selecting evidence passages?'' \emph{Determines whether the retrieval judge emphasizes recall-oriented or precision-oriented families and sets its direction of goodness.}
\item \textbf{Answer Generation \& Formatting:} ``Should an answer that abstains (``insufficient evidence'') count as a success or a failure?'' \emph{Sets the direction of goodness for the abstention/refusal-rate metric: counted as success, the rate should be high; counted as failure, low.}
\item \textbf{Agent Execution Loop:} ``Is a high tool-retry / re-search rate expected thoroughness or a sign of degradation?'' \emph{Sets the direction of goodness and threshold for the tool-retry-rate metric.}
\item \textbf{Metadata Query Orchestrator:} ``When a metadata source (CrossRef, Semantic Scholar) is unavailable, is falling back to another source acceptable or should it be flagged?'' \emph{Decides whether the cross-source fallback-rate is monitored as healthy behavior or as a failure signal.}
\item \textbf{PDF reader plugins:} ``Which parser (Docling, PyMuPDF, PyPDF) carries production traffic, and which are fallbacks?'' \emph{Scopes the parse-failure-rate metric to the production parser rather than penalizing fallbacks.}
\item \textbf{Nemotron reader:} ``Is the Nemotron API the default page parser or an optional high-accuracy path?'' \emph{Determines whether its availability and latency metrics are treated as production-critical.}
\item \textbf{Vector Store (Embedding Index):} ``What index staleness is acceptable before results are considered degraded?'' \emph{Sets the threshold for the index-freshness monitor.}
\end{itemize}

\paragraph{How we built the ground truth.}
Without looking at Litmus's output, we listed components from three sources we already have: the rule classes in \texttt{override\_rules.json}; one component per distinct routing label (override, disaggregation, RAG, orchestrator, retrieval, operational, general); and the infrastructure dependencies named in the source---then one practitioner adjusted the list and confirmed boundaries. A node is a true positive if it matches a ground-truth component at the same granularity; spurious nodes are false positives, unmatched components false negatives. The AI-logic variant keeps only components with model calls or routing logic.

\begin{table}[t]
\centering
\small
\begin{tabular}{lc}
\toprule
\textbf{Fidelity measure} & \textbf{Value} \\
\midrule
Component nodes recovered            & 30 \\
Subsystems spanned                   & 7 \\
Ground-truth source modules          & 32 \\
Data-flow edges verified             & 65\% \\
Critic overall confidence            & 0.78 \\
Critic lowest-node confidence        & 0.65 \\
Phantom nodes pruned                 & 0 \\
Missed components flagged            & 3 \\
Coverage P / R / F1 (all)            & 1.00 / 1.00 / 1.00 \\
Coverage P / R / F1 (AI-logic)       & 1.00 / 1.00 / 1.00 \\
Per-module F1 (strict)               & 0.64 \\
\bottomrule
\end{tabular}
\caption{Architecture-reconstruction fidelity for the scientific-QA pipeline (PaperQA2). The seven subsystems are PDF reader plugins, agent orchestration, metadata clients, the core QA engine, the Nemotron reader, contributed sources, and tests. Strict per-module F1 (0.64) is lower than coverage F1 because Litmus clusters the 32 source modules into 30 functional nodes as shown in Figure \ref{fig:archpqa} rather than emitting one node per file. All values are recorded from the persisted graph and critic annotations (Appendix~\ref{app:arch}).}
\label{tab:archfidelity-pqa}
\end{table}

\paragraph{Fidelity vs.\ validity pairing.}
The $n{=}7$ pooled points in Section~\ref{sec:fidelityvalidity} are the scoreable metrics (non-null criterion validity) across the three pipelines, each paired with its host-subsystem mean critic confidence. Component-level $n{=}3$ is one point per pipeline. For P2 the host is \texttt{core-qa-engine} (five nodes, mean $0.83$); the $0.78$ above is the mean over all 30 nodes.

\paragraph{How the numbers were recorded.}
The graph and critic annotations are persisted per solution in the system's own store, so these figures are recomputable in a deployment without new model calls; the client artifacts themselves are proprietary and are not released (\hyperref[sec:ethics]{Ethical Considerations}). The queries below are the recording path, not a public reproduction recipe:
{\footnotesize
\begin{verbatim}
# dump latest graph + critic rows
sqlite3 litmus.db \
  "SELECT graph_json
   FROM architecture_graphs
   ORDER BY version DESC
   LIMIT 1;" > graph.json
sqlite3 litmus.db \
  "SELECT critic_annotations
   FROM architecture_graphs
   ORDER BY version DESC
   LIMIT 1;" > critic.json

# edge verification counts
jq '[.edges[].verification]
    | group_by(.)
    | map({(.[0]): length})
    | add' graph.json

# critic confidence + missing nodes
jq '.overallConfidence,
    [.annotations[]
     | select(.confidence < 0.3)],
    .missingNodes' critic.json
\end{verbatim}
}
\section{Deterministic Algorithms and LLM Prompts by Component}
\label{app:impl}

This appendix documents each major Litmus component at the implementation level. For every phase we separate the two kinds of stages Litmus interleaves: \emph{deterministic} stages (static analysis, graph construction, verification, rule-based selection), and \emph{LLM} stages (synthesis, classification, critique, design) which are given as the verbatim system and user prompts issued to the model. Prompts are reproduced as sent, with three cosmetic normalizations for typesetting only: (i) runtime-interpolated values appear as \texttt{\{placeholder\}}; (ii) non-ASCII decorations in the original strings (box-drawing rules, bullets, arrows, $\leq$/$\geq$) are mapped to ASCII; and (iii) embedded JSON schemas, which are generated programmatically from Zod definitions, are summarized rather than reprinted. Unless noted otherwise, every LLM call targets a Claude Opus-class model through an OpenAI-compatible gateway and requests structured JSON output; per-call temperature and token budget are stated with each prompt.

\subsection{Architecture Reconstruction}
\label{app:impl-arch}

A deterministic front end scans the repository, extracts tree-sitter structure, builds a symbol/call graph, and computes centrality. An LLM then synthesizes the component graph (single-pass, or the multi-pass cluster/inflate/stitch variant); deterministic passes verify every edge against the call graph, assign node kinds and edge types, and repair connectivity; finally an adversarial LLM critic re-reads the source.

\paragraph{LLM prompts (architecture).} The synthesizer, the multi-pass passes, and the critic share the analyst persona below. All synthesis calls use temperature $0.1$; the critic uses temperature $0$. Token budget is large ($\sim$128k) to fit whole-repository context.

\promptlabel{Shared analyst system prompt (\texttt{buildArchitectureSystemPrompt}); used by the synthesizer and the inflate pass.}
\begin{lstlisting}[style=prompt]
You are an AI systems analyst for evaluation engineering.

Your task is to reconstruct the true execution architecture of this codebase from code evidence, then annotate that architecture so an AI engineer can understand:
1. how the system actually works,
2. which components are AI-core versus supporting or operational,
3. which components should receive metrics first.

PRIORITY ORDER:
1. Execution truth
2. AI semantics
3. Evaluability and observability
4. Framework alignment

EXECUTION TRUTH:
- Infer the real runtime/dataflow from code, not from names.
- Identify triggers, transformations, external calls, storage boundaries, branching logic, outputs, and feedback paths.
- Prefer fewer truthful nodes over many shallow nodes.
- Collapse helpers into parent nodes unless a helper has a distinct evaluation or observability surface.

AI SEMANTICS:
Treat the following as first-class architectural behavior when present:
- prompt_construction
- model_call
- retrieval
- reranking
- tool_use
- memory
- guardrail
- judge
- fallback
- post_processing

SYSTEM ROLE:
- ai_core: directly shapes semantic behavior, ranking, judgments, tool choice, or final answer quality
- supporting: materially supports or constrains AI quality, freshness, latency, or traceability
- operational: broader system health, reliability, observability, or governance surfaces

METRIC PRIORITY:
- primary: components that directly shape semantic behavior or user-visible AI quality
- secondary: components that materially support, constrain, or distort AI quality
- tertiary: operational components whose metrics are useful but not first-order for AI evaluation
- Do not use framework alignment to determine metric priority.

FRAMEWORK OVERLAY:
- After reconstructing the real graph, annotate each node with whether it is strongly aligned to a framework stage, partially aligned, or adjacent.
- A node is adjacent if it materially affects AI behavior, quality, latency, reliability, or observability, even if it is not a canonical framework component.
- Do not force nodes into the framework based on naming alone.

EVALUATION OVERLAY:
For each node, determine:
- whether it is offline-evaluable
- whether it is runtime-observable
- whether quality can be measured directly, requires judge-style evaluation, or only has runtime proxies
- what user-facing failure happens if the node is wrong

NODE RULES:
- Each node must map to real files or functions.
- Each node must describe its primary runtime responsibility and downstream effect.
- Keep descriptions concise and return compact JSON.
- Prefer evaluability-relevant decomposition over generic software decomposition.
- Preserve adjacent framework components if they still matter for evaluation.

EDGE RULES:
- Edges should represent actual execution or control transitions.
- Label edges when the transition matters for quality, failure, fallback, or observability.
- Include feedback loops when retries, evaluations, or guardrails influence later behavior.

DO NOT:
- optimize for a pretty diagram over a truthful one
- flatten AI behavior into generic "processing"
- exclude adjacent components that still need evaluation
- assume the framework is the system; it is only an overlay
- mark a node as strongly aligned without code evidence
\end{lstlisting}

\promptlabel{Cluster-pass system prompt (multi-pass synthesis, temperature 0.1).}
\begin{lstlisting}[style=prompt]
You are a codebase clustering engine. Group source files into 3-8 named subsystems based on their function call relationships and framework roles.

RULES:
1. Every source file must belong to exactly one subsystem.
2. Minimize cross-subsystem call edges (high cohesion, low coupling).
3. Name subsystems by what they DO, not what they ARE (e.g. "Document Retrieval" not "Module A").
4. Infrastructure nodes (databases, caches, queues, vector stores) form their own subsystem only if they have 3+ files; otherwise attach to their primary consumer.
5. Entry points (API routes, Lambda handlers, CLI) group together unless they serve clearly different domains.
6. Keep DISTINCT business/domain modules separate. If two groups of files implement different domain pipelines -- different domain vocabulary, inputs/outputs, or end-to-end flow -- they MUST be separate subsystems. Never collapse them into one bucket.
7. Do NOT emit catch-all subsystems with vague names like "Domain Logic", "Core Logic", "Business Logic", "Processing", or "Miscellaneous" that lump unrelated domains together. Split such a group into its constituent domains, each named for what it does.
8. Choose the number of subsystems (3-8) that reflects the codebase's actual distinct domains and infrastructure -- do not over-merge to hit a smaller count.

Return ONLY valid JSON matching the schema. No markdown. {JSON schema}
\end{lstlisting}

\promptlabel{Stitch-pass system prompt (connects nodes across subsystems).}
\begin{lstlisting}[style=prompt]
You are a cross-subsystem edge generator. Connect architecture nodes across different subsystems to ensure the graph is fully connected.

RULES:
1. Only create edges between nodes in DIFFERENT subsystems.
2. For every edge, populate evidenceCalls with caller->callee symbol pairs when symbol evidence is available. If no symbol evidence exists, use descriptive evidence like "data_flow: subsystemA.output -> subsystemB.input".
3. Do NOT duplicate edges that already exist within subsystems.
4. Use the CROSS-SUBSYSTEM SYMBOL EDGES section (if present) to identify which files call across subsystem boundaries, then connect the architecture nodes that own those files.
5. CONNECTIVITY IS THE TOP PRIORITY: Every subsystem MUST have at least one edge connecting it to another subsystem. No subsystem may be isolated.
6. If symbol evidence is absent for a subsystem, infer edges from data flow patterns (e.g., orchestration calls processing, entry points feed pipelines, data subsystems serve compute subsystems).
7. Generate enough edges to make the graph navigable -- aim for at least one edge per subsystem pair that has a logical data flow relationship.

Return ONLY valid JSON matching the schema. No markdown. {JSON schema}
\end{lstlisting}

\promptlabel{Architecture critic system prompt (temperature 0).}
\begin{lstlisting}[style=prompt]
You are an architecture critic. Your job is to review an architecture graph produced by another AI agent and verify it against the actual source code.

CHECK FOR:
1. PHANTOM NODES: Nodes that claim source files that don't exist
2. MISSED NODES: High-importance files not represented by any node
3. WRONG EDGES: Claimed data flow that doesn't match import structure
4. MISCLASSIFIED ROLES: Nodes labeled as "ai_core" that don't use AI libraries
5. DUPLICATE NODES: Two nodes representing the same component

For each node, provide:
- confidence (0-1): how confident you are this node is accurate
- issues: list of problems found (empty if none)
- suggestions: list of improvements

Be adversarial. Assume the synthesizer made mistakes. Verify claims against the code.

Return ONLY valid JSON matching the schema. {JSON schema}
\end{lstlisting}

\promptlabel{Architecture critic user prompt.}
\begin{lstlisting}[style=prompt]
Review this architecture graph for accuracy:

GRAPH:
{graph as JSON}

SOURCE CODE EXCERPTS:
{for each node, up to 2 source files (<=4000 chars each), or FILE NOT FOUND}

FILE MANIFEST (for detecting missed nodes):
{path [centrality=score] per file}
\end{lstlisting}

\subsection{Pattern-Based Metric Search Space}
\label{app:impl-pattern}

Classification narrows the metric search space by mapping each component to an evaluation-pattern taxonomy: six \emph{pipeline stages} (data acquisition, data transformation, knowledge base, AI core, assurance, continuous improvement) and five \emph{cross-cutting strategies} (RAG, evidence matching, prompt-chaining, agentic, LLM-as-judge). A holistic LLM classifier sees the whole graph and emits per-component stage, cross-cutting strategies, and a single dominant pipeline pattern; an adversarial LLM validator checks the result; deterministic helpers provide a rule-based fallback.

\paragraph{LLM prompts (classification).} The classifier runs at temperature $0.1$; the validator at temperature $0$. Both receive the taxonomy catalog (stage definitions and detection heuristics) inlined as JSON in place of \texttt{\{catalog\}}.

\promptlabel{Holistic classifier system prompt.}
\begin{lstlisting}[style=prompt]
You are a holistic architecture classifier for AI evaluation engineering.

Your task is to analyze the FULL architecture graph and:
1. Classify each component into a pattern stage from the framework catalog.
2. Detect cross-cutting strategies that span multiple components.

Unlike a per-component classifier, you can see the entire graph topology and detect graph-level patterns.

{catalog: PATTERN STAGES and CROSS-CUTTING STRATEGIES as JSON}

CROSS-CUTTING DETECTION RULES:

RAG (Retrieval-Augmented Generation):
- Retriever component feeding into a generator component with shared context
- Vector store query followed by LLM call
- Context injection into prompt from retrieved documents

Prompt Chaining:
- Sequential LLM calls where output of one feeds into input of next
- Multi-step pipeline with intermediate LLM processing
- Chain or workflow orchestration across multiple prompts

Agentic:
- Tool-use loops and decision nodes in the graph
- ReAct loops or iterative decision-making
- Autonomous multi-step execution with branching

Evidence Matching:
- Components that compare/match evidence against criteria
- Matching or reconciliation logic between two data sources
- Verification of assertions against supporting documents

LLM-as-Judge (judge):
- Component evaluates, scores, ranks, or compares ANOTHER LLM's or AI model's OUTPUT
  (not domain objects like risks, accounts, or documents)
- Rubric-based scoring with explicit criteria and a numeric scale applied to
  model-generated text or structured output
- Pairwise preference / A-vs-B comparison of two model responses
- Calibration / agreement testing against human-labeled data ON MODEL OUTPUTS
- Outputs structured verdicts (pass/fail, score, reasoning) about another
  component's AI-generated output

NOT judge if:
- The component scores, classifies, or assesses DOMAIN OBJECTS (risks, accounts,
  documents, transactions) even if it uses an LLM to do so
- The component is a domain classifier, ranker, or assessor that happens to
  produce scores -- that is "prompt_chain" or "rag", not "judge"

PRIMARY PIPELINE PATTERN (pipelinePattern):
For EACH component, you MUST also emit a single dominant pipelinePattern label
from this fixed set:
  rag | agentic | judge | prompt_chain | evidence_match | none

Decision rules (apply in order -- pick the FIRST that matches):
1. If the component is itself an LLM-as-Judge / scorer / grader
   THAT EVALUATES ANOTHER AI MODEL'S OUTPUT (not domain objects) -> "judge"
2. Else if the component is part of an agent loop (tool-calling, ReAct,
   planning) -> "agentic"
3. Else if the component retrieves external context and feeds it to an LLM
   (or is the retriever in a clear retriever->generator pipeline) -> "rag"
4. Else if the component compares/reconciles evidence between two sources
   -> "evidence_match"
5. Else if the component is one step in a sequential multi-LLM chain
   -> "prompt_chain"
6. Else (deterministic ETL, schema validation, infra glue, plain inference
   with no retrieval, etc.) -> "none"

A component's pipelinePattern is INDEPENDENT of its patternStage -- e.g. a RAG
retriever has pipelinePattern="rag" AND patternStage="knowledge_base". Do NOT
return "none" just because the component is non-LLM; only pick "none" when no
LLM-pipeline pattern applies.

CLASSIFICATION RULES:
- Set patternStage to null for components that do not map to any framework stage.
- pipelinePattern is REQUIRED for every component (use "none" if no pattern fits).
- confidence should reflect how certain you are (0.0 to 1.0).
- crossCuttingStrategies should list all strategies the component participates in (multi-select; can include "judge").
- reasoning should be a concise explanation of your classification decision.
- detectedPatterns should list all graph-level patterns you detect with the involved component IDs.

Return ONLY valid JSON matching this schema. No markdown, no explanation. {JSON schema}
\end{lstlisting}

\promptlabel{Classification validator user prompt (source digests loaded only for items with confidence $<0.8$).}
\begin{lstlisting}[style=prompt]
Validate the following classifications against the architecture graph.

CLASSIFICATIONS:
{classifications as JSON}

DETECTED PATTERNS:
{detectedPatterns as JSON}

ARCHITECTURE GRAPH:
{graph as JSON}

SOURCE CODE DIGESTS (for low-confidence items)
\end{lstlisting}

\subsection{Interrogation: Turning Assumptions into Constraints}
\label{app:impl-interrogation}

Interrogation runs two channels. \emph{Clarification questions} to the practitioner are generated deterministically from AST features and detected code patterns; each is emitted with a \emph{metric-impact} statement and capped at three per component. \emph{Validity questions} are posed by an adversarial LLM critic that issues a per-metric verdict (pass / reject / flag) after a structured four-axis review (pipeline fit, validity conditions, data \& signal assumptions, direction of goodness). Confirmed answers become hard constraints: a deterministic policy step rejects metrics whose threshold contradicts a confirmed direction-of-goodness answer and downgrades unverified-direction rejections to soft flags when no fact covers them.

\paragraph{LLM prompts (validity critic).} The critic reviews all metrics across components in batches (temperature $0$). The full system prompt is reproduced below; ASCII rules replace the original box-drawing separators.

\promptlabel{Adversarial metric critic system prompt.}
\begin{lstlisting}[style=prompt]
You are an adversarial metric critic for AI evaluation engineering. You review ALL metrics across ALL components at once to enforce quality, consistency, and deduplication.

Your job is to issue a verdict for EVERY metric: PASS, REJECT, or FLAG -- with a structured, multi-axis review that tells the metric designer exactly how to fix it.

=======================
FOUR-AXIS REVIEW -- run this mentally for every metric before picking a verdict.
=======================

1. PIPELINE FIT -- does this metric make logical sense for where this component sits?
   - What does this component actually consume from upstream, and what does it emit downstream?
   - Can this component genuinely influence what the metric measures, or does the real signal
     live in a different component entirely? (e.g. a retriever being scored on generation quality)
   - Is the measurement captured at a point where the data it needs is already present?
   - Does the metric generalise to ANY component of the same type, or does it bite into what
     THIS component uniquely does? Generic-fit metrics are a REJECT, not a FLAG.

2. VALIDITY CONDITIONS -- under what specific conditions does this metric produce signal?
   - Enumerate the runtime conditions that must hold for the number to be meaningful
     (e.g. "non-empty retrieval result", "English query", "user session has prior turn",
      "source doc contains the entity being cited").
   - Identify cases where the metric will return a value but MEAN NOTHING (cache hits,
     fallback paths, empty inputs falling through, early exits). If those failure modes
     dominate real traffic, this is a REJECT.
   - Are the valid conditions realistic given how the component is exercised in production,
     or does the metric only fire on a narrow sliver of requests?

3. DATA & SIGNAL ASSUMPTIONS -- what must be true of the system for this measurement to work?
   - List every artefact/signal the measurement depends on: labels, ground truth, judge
     model, traced IDs, structured logs, schemas, embedding store, golden datasets, cost
     tracking, user feedback signal, etc.
   - Cross-check the source digest: do those artefacts actually exist in THIS codebase? If
     the metric silently assumes something that isn't there (e.g. "compare to ground truth"
     when no labelled set exists), that is a REJECT.
   - Surface any assumption that would be wrong without the reader knowing -- silent
     assumptions are the most dangerous kind.
   - DIRECTION OF GOODNESS -- every metric encodes a directional prior via its threshold
     operator (> says higher is better, < says lower is better). Extract that prior
     as a one-sentence claim. Then search the source digest AND any user-supplied hint
     for textual evidence that anchors it. If an anchor exists, quote/paraphrase it as
     evidence and mark verdict: "verified". If no anchor exists -- the direction is
     the designer's generic prior, not the codebase's intent -- mark verdict: "unverified"
     with evidence: null. Emit a directionAssumption block on EVERY verdict (pass /
     flag / reject). Pipeline-flow metrics (tier-hit-rate, fallback-depth, cache-hit-rate,
     retry-count, refusal-rate) are the high-risk family -- wrong direction here inverts
     production alerts.

   CONFIRMED FACTS RULES (M7):
   a) If the component header includes a CONFIRMED FACTS block, those answers are
      GROUND TRUTH for direction-of-goodness. A metric whose threshold contradicts
      a confirmed direction answer is a REJECT -- cite the confirmed fact as evidence.
      When a confirmed fact supports the direction, set evidence to the confirmed
      answer text and verdict: "verified".
   b) If NO confirmed fact covers direction for this component, do NOT infer direction
      from the source digest alone. Mark verdict: "unverified" -- soft warn, NOT
      REJECT. The user has not confirmed direction, so assumptions are flagged, never
      enforced.

4. QUALITY CHECKS -- apply the REJECT / FLAG criteria below.

=======================
REJECT if:
=======================
- Metric fails PIPELINE FIT: it measures something this component can't actually influence, or it fits the role generically rather than this component's actual behaviour.
- Metric fails VALIDITY CONDITIONS: the conditions for it to mean anything rarely hold, or the metric will silently report meaningless values on common paths.
- Metric fails DATA & SIGNAL ASSUMPTIONS: it relies on artefacts/signals that don't exist in this codebase and can't be practically collected.
- Metric measures implementation, not outcomes (e.g., "function call count" instead of "task completion rate").
- Threshold is arbitrary with no justification (e.g., "> 0.8" with no rationale).
- Non-deterministic component lacks an LLM-as-Judge metric.
- LLM-as-Judge metric is missing a scoring rubric or Chain-of-Thought (CoT) instruction.
- 5 Metric Rule violated: a component has more than 5 metrics.
- TOO GENERIC: The metric name and description could apply to any component of the same type. A metric named "Response Quality" or "Output Accuracy" that doesn't reference what THIS specific component does is too generic and MUST be rejected. Ask: "Would this metric name make sense for a completely different component?" If yes, it is too generic.
- NOT MEASURABLE: The computationApproach references data, signals, or capabilities that are not available in the codebase or cannot be practically collected.
- DOESN'T MATCH CODE: The metric measures something the component doesn't actually do. If the source files show the component is a simple pass-through or formatter, metrics about "semantic quality" or "reasoning depth" are wrong and must be rejected.
- ORTHOGONAL COMPOSITE: The metric combines unrelated dimensions into a composite (e.g., "0.5*Accuracy + 0.5*Latency"). Sub-metrics in a composite must measure the SAME quality from different angles. Combining orthogonal concerns is always wrong.
- UNVERIFIED DIRECTION ON CASCADE FAMILY: For pipeline-flow metrics whose name or intent matches the cascade family (tier-hit-rate, fallback-depth, fallback-rate, cache-hit-rate, retry-count, retry-success-rate, refusal-rate, escalation-rate, RAG-hit-rate, and similar), an unverified direction-of-goodness assumption is REJECT (not FLAG). The failure mode here is inverted production alerts; the designer must either ground the direction in the source digest or take it from the user-supplied hint.
- VAGUE UMBRELLA JUDGE: When a component has an INTERNAL STRUCTURE block listing tiers/sub-modules (or its description/edge types reveal multi-tier structure), an LLM-as-Judge metric that covers the whole module vaguely (e.g. "X_semantic_correctness", "X_output_quality", "X_grounding") while one or more precise tier-level judges already exist in the same metric set is REJECT. The vague metric adds no signal the precise ones don't cover. Each LLM-as-Judge must target a specific {LLM} tier's unique failure mode. If the component has N tiers marked {LLM}, expect up to N precise judges -- not N-1 precise + 1 vague umbrella, and not extra judges for {rule-based} tiers.
- RETRIEVAL MEASURED BY LLM-JUDGE: An LLM-as-Judge metric (scoringRubric != null) whose name/intent is a retrieval-quality measure -- context/chunk relevance, retrieval precision, recall@k, precision@k, MRR, NDCG, context precision -- is REJECT. These are computable by embedding similarity or framework scorers (RAGAS/DeepEval), so an LLM judge wastes compute. suggestedFix: convert to a framework-based metric (scoringRubric=null, computationApproach using the relevant scorer). This does NOT apply to genuinely semantic judges (faithfulness, groundedness, correctness, coherence) -- those legitimately need an LLM judge.
- JUDGE ONLY LLM TIERS: When the INTERNAL STRUCTURE block is present, an LLM-as-Judge metric (scoringRubric != null) that targets a tier marked {rule-based} -- a deterministic rule/keyword/exact-match step or the top-level orchestrator/router -- is REJECT. A judge is only meaningful where an LLM produces the output; a {rule-based} tier has no LLM output to judge and is covered by deterministic metrics. Match the judge to its tier by name/intent; if a judge plainly evaluates a {rule-based} tier, REJECT it. suggestedFix: drop the judge, or replace it with a deterministic check (scoringRubric=null) if a measurable failure mode exists. This does NOT reject judges on {LLM} tiers.
- INCOMPLETE COST_PER_QUALITY TRIPLE-EMIT: For category="cost_per_quality" metrics, the designer must emit a langfuseConfig bundle containing all three of: (a) costScorer reading usage.totalCost / usage.cost / sum_costDetails, (b) qualityScorer with pairedMetricName referencing another metric on THIS component (must be a quality / groundedness / safety metric, NOT another cost metric), (c) derivedMetric with a cost-per-pass formula. Missing any of the three, or a qualityScorer pairing to a non-existent metric on the same component, or a qualityScorer pairing to another cost_per_quality metric, is REJECT -- a cost number alone is unactionable in production.

=======================
DETERMINISTIC CATALOG METRICS -- read this carefully.
=======================
Component headers list `Deterministic metrics` -- these are pattern-keyed templates (error_rate_per_class, retry_count_p95, cache_hit_rate, llm_token_p95, etc.) instantiated from a regex match on the source digest, NOT designed by the LLM. They intentionally generalise across components.
- Do NOT REJECT for TOO_GENERIC: their names are template names by design.
- Do NOT REJECT for DOESN'T_MATCH_CODE solely because the metric is broad: the regex detector already verified the pattern is present.
- DO still apply VALIDITY CONDITIONS, DATA ASSUMPTIONS, and DIRECTION OF GOODNESS checks.
- DO emit FLAG if the template's threshold is clearly wrong for THIS component.
- These metrics carry type: "monitoring" and scoringRubric: null by design; do NOT REJECT for missing rubric.

=======================
DETERMINISTIC COVERAGE.
=======================
Component headers also list `Detected code patterns` (has_try_except, has_retry_loop, has_cascade, parses_json, calls_external_api, calls_llm, has_cache, has_db_write, streams_response, has_queue, has_circuit_breaker). For each detected pattern, verify that the component's metric set contains at least one metric whose intent matches that pattern's failure mode. If a pattern is detected but uncovered, add a crossComponentIssues entry with issue: "missing_deterministic_coverage: <pattern_id>" and a one-line suggestion.

=======================
PER-LLM-TIER COVERAGE -- for components with an INTERNAL STRUCTURE block.
=======================
Coverage applies ONLY to tiers marked {LLM}. Every {LLM} tier should have at least one metric targeting its unique semantic failure mode. Tiers marked {rule-based} and the orchestrator do NOT need a metric here. For each {LLM} tier with NO matching metric, add a crossComponentIssues entry with issue: "missing_tier_coverage: <tier label>".

=======================
FLAG if:
=======================
- Two components have metrics measuring the same thing (cross-component duplicate).
- Composite metric weights don't sum to 1.0.
- An eval metric has no monitoring counterpart (and vice versa when appropriate).
- A validity condition or data assumption is plausible but NOT verifiable from the source digest.
- UNVERIFIED DIRECTION (non-cascade): a direction-of-goodness claim with no anchor in digest or hint, for a metric outside the cascade family above, is FLAG.

=======================
VERDICT DISCIPLINE.
=======================
- "pass" -- correct across ALL FOUR axes; set review to null. The pipeline SKIPS redesign for components whose metrics all pass, so false flags cost an entire redesign cycle.
- "flag" -- conceptually right but has a surfaced concern; emit a FULL review block.
- "reject" -- fails at least one axis; emit a FULL review, concise reasoning (<= 3 sentences), and a specific, actionable suggestedFix.

OUTPUT CONTRACT: for every verdict include metricName, componentId, verdict, reasoning, review (null for pass, else {pipelineFit, validityConditions[], dataAssumptions[], concerns[]}), suggestedFix (reject/flag), and directionAssumption {claim, evidence|null, verdict: verified|unverified} on EVERY verdict. Report cross-component issues in crossComponentIssues and a short overallAssessment.

Be adversarial on quality but disciplined on verdicts. Return ONLY valid JSON matching the schema. No markdown, no explanation. {JSON schema}
\end{lstlisting}

\subsection{Metric Design}
\label{app:impl-metric}

Metric design is conditioned on the reconstructed architecture, the component classifications, and the confirmed facts. Deterministic templates (counters, rates, percentiles keyed on detected patterns; framework and node-kind metrics; pipeline-pattern presets) are instantiated first and merged with the LLM-designed metrics, which target semantic-quality gaps the deterministic ones cannot reach. Litmus supports three metric types: \emph{deterministic checks} (\texttt{scoringRubric}=null), \emph{LLM-as-judge} metrics (a 5-point rubric plus a chain-of-thought instruction), and \emph{composite} metrics (weighted sub-metrics measuring one quality from several angles). Selection is bounded by the five-metric rule; the redesign loop re-invokes the designer on rejected/flagged LLM metrics and uncovered LLM tiers.

\paragraph{LLM prompts (designer).} The designer runs at temperature $0.2$ for a single component at a time. The system prompt embeds the evaluation methodology and worked examples; dynamic blocks (pattern preset, enabled categories, export targets, stage and cross-cutting catalogs) are inlined where marked.

\promptlabel{Metric designer system prompt (hard constraints; \texttt{\{...\}} blocks inlined at runtime).}
\begin{lstlisting}[style=prompt]
You are a metric designer for AI evaluation engineering. You design high-signal metrics for a SINGLE component.

HARD CONSTRAINTS:
1. MAXIMUM {maxMetrics} LLM-JUDGE METRICS this call may emit. Deterministic counters/rates/percentiles already arrive via the PRE-CHOSEN METRICS block (user prompt) and are merged automatically -- do NOT duplicate them. Quality over quantity -- fewer, deeper metrics.
2. DOMAIN-SPECIFIC: Each metric must reflect what this component ACTUALLY DOES.
   A retriever needs retrieval precision -- not generic "accuracy".
   A prompt builder needs instruction completeness -- not "latency".
   An agent orchestrator needs convergence score -- not "response quality".
   Ask: "What are the specific ways this component can fail?"
3. ONE CRITERION PER METRIC: Never combine unrelated dimensions. Relevancy and clarity are separate metrics, not one.
4. RIGHT TOOL: Use code-based computation for anything measurable (format, thresholds, regex, counts, latency). Use LLM-as-Judge ONLY for semantic quality that cannot be computed deterministically.
   RETRIEVAL METRICS ARE NOT LLM-JUDGE: Retrieval relevance, precision@k, recall@k, MRR, NDCG, and context relevancy are computable via embedding similarity or framework scorers (RAGAS, DeepEval, etc.) -- NOT via LLM-as-Judge. Set scoringRubric to null for these and describe the framework-based computationApproach. Reserve LLM-as-Judge for truly semantic assessments like answer faithfulness, grounding, coherence, or domain-specific correctness.
5. NO REDUNDANCY: If two candidate metrics are correlated, keep only the one with higher signal.
6. Non-deterministic components MUST have at least one LLM-as-Judge metric with a 3-point scoring rubric and CoT.
7. Every threshold MUST include a justification grounded in the component's risk surface and blast radius -- no arbitrary numbers.
8. COMPOSITE WHEN NATURAL: If a quality dimension has genuinely distinct sub-dimensions (e.g. traceability = citation coverage + link strength + evidence breadth), design a composite metric with weighted sub-metrics and a formula. Do NOT force composite structure when a single measurement suffices -- set both to null.
9. Monitoring metrics should have a null scoringRubric.
10. COMPOSITE ANTI-PATTERN: NEVER combine orthogonal dimensions into a composite. Accuracy + Latency is NOT a valid composite. A composite is ONLY valid when sub-metrics measure the SAME quality from different angles (e.g. retrieval quality = precision + recall + MRR). If in doubt, use separate single-dimension metrics.
11. GRANULAR JUDGES -- NO VAGUE UMBRELLA METRICS: When a component has internal sub-modules, tiers, or distinct LLM stages, design one metric per unique sub-concern -- each targeting a specific tier's distinct function. NEVER emit a broad module-level metric (e.g. "X_grounding", "X_semantic_correctness", "X_output_quality") that spans multiple tiers. Instead, emit tier-specific metrics: e.g. "tier1_retrieval_precision", "tier2_processing_correctness", "tier3_output_faithfulness". Each metric must name the tier it targets.
12. COST_PER_QUALITY TRIPLE-EMIT: For category="cost_per_quality" you MUST emit a langfuseConfig containing ALL THREE of: (a) costScorer reading usage.totalCost / usage.cost / sum_costDetails, (b) qualityScorer referencing one of the OTHER metrics in this same component's set via pairedMetricName (a quality/groundedness/safety metric -- never another cost metric), (c) derivedMetric whose formula divides cost by quality-pass count. Cost without a paired quality scorer is REJECT. Non-cost_per_quality metrics MUST omit langfuseConfig.

DETERMINISTIC STATUS: {componentDeterministic}
(If "no": this component is NON-DETERMINISTIC -- at least one evaluation metric MUST use LLM-as-Judge with a 3-point rubric. If "partial": consider whether LLM-as-Judge metrics are needed for non-deterministic aspects.)

GOOD vs BAD METRICS -- learn from these examples:
BAD: "Response Quality Score" for an LLM inference component
  Why bad: Generic name that could apply to ANY LLM component.
BAD: "Processing Accuracy" for a chunking component
  Why bad: "Accuracy" is vague -- chunk boundaries? content preservation? overlap?
GOOD: "Citation Grounding Rate" for a document generation component
  Why good: Specific failure mode (claims without source backing).
GOOD: "Chunk Boundary Coherence" for a semantic chunking component
  Why good: Measures whether chunk splits preserve semantic completeness.
BAD: "rag_retrieval_relevance" as an LLM-as-Judge metric with a 3-point rubric
  Why bad: Retrieval relevance is measurable by embedding similarity / RAGAS / MRR.
GOOD: "rag_retrieval_relevance" as a framework-based metric (scoringRubric=null, RAGAS context_precision).
BAD: a multi-tier module gets a vague umbrella judge plus only one tier-specific metric.
GOOD: the same module gets one precise metric per tier (framework-based for retrieval, LLM-judge for semantic tiers).

EXEMPLAR -- a well-designed composite metric:
Supporting-Document Traceability Index (SDTI) for a document generation component:
- Composite 0-1 score: 0.4*CitationCoverage + 0.3*AvgLinkStrength + 0.3*EvidenceBreadth
- CC: fraction of sentences with an explicit citation or cosine-sim >= 0.78 to source chunk
- ALS: mean cosine similarity of supported sentences to their evidence
- EB: unique source docs cited / total available source docs
- Threshold: >= 0.7, justified by audit compliance requirements
- Note: all three sub-metrics measure traceability from different angles.

EXEMPLAR -- a well-formed cost_per_quality TRIPLE-EMIT:
Cost Per Faithful Answer for an LLM-as-Judge component:
- threshold: { value:"0.05", operator:"lte", justification:"per-answer cost above $0.05 breaks unit economics" }
- langfuseConfig:
    costScorer: { name:"trace_cost_usd", source:"usage.totalCost", aggregation:"sum" }
    qualityScorer: { name:"answer_faithfulness_pass", type:"llm_judge", pairedMetricName:"answer_faithfulness", passCondition:"score >= 2" }
    derivedMetric: { name:"cost_per_faithful_answer", formula:"trace_cost_usd / max(answer_faithfulness_pass, 1)" }

{PIPELINE PATTERN block}
{METRIC CATEGORIES block}
{EXPORT TARGETS block}
{FALLBACK STAGE CATALOG}
{CROSS-CUTTING STRATEGY METRICS}

EVALUATION METHODOLOGY:
5-POINT SCORING SCALE (for LLM-as-Judge metrics):
- 5 (Excellent): Fully meets all criteria with no issues
- 4 (Good): Meets criteria with only minor issues
- 3 (Acceptable): Partially meets criteria with some issues
- 2 (Poor): Largely fails to meet criteria with major issues
- 1 (Very Poor): Fails to meet criteria entirely
CHAIN-OF-THOUGHT (CoT) INSTRUCTION: every LLM-as-Judge metric must tell the judge to
(1) explain its reasoning step by step, (2) consider both strengths and weaknesses,
(3) then assign a score based on the rubric.
ANTI-BIAS INSTRUCTIONS:
- Do not favor longer or more verbose outputs
- Judge based on criteria, not style or format
- Consider the specific use case and context

OUTPUT TAGS (REQUIRED for each metric):
- category: one of {quality | groundedness | safety | drift | reliability | latency | cost_per_quality | tool_use | judge_alignment} (must be one the user enabled).
- applicableTargets: array of {in_process | langfuse | dashboard | alert | ci_gate} (only targets the user enabled).

Return ONLY valid JSON matching this schema. No markdown, no explanation. {JSON schema}
\end{lstlisting}

\subsection{Traceability and Export}
\label{app:impl-export}

The final phase preserves rationale and produces runnable artifacts. An LLM traceability mapper links each evaluation metric to its monitoring counterpart, flags monitoring gaps, and proposes alert configurations and instrumentation points. Deterministic steps enforce justification quality, standardize externally detected metrics into the common return schema, and emit code.

\paragraph{LLM prompt (traceability).} The mapper runs per component at temperature $0.1$, receiving the approved evaluation and monitoring metrics separately.

\promptlabel{Traceability mapper system prompt.}
\begin{lstlisting}[style=prompt]
You are a traceability mapper for AI evaluation engineering. You map evaluation metrics to their monitoring counterparts, identify monitoring gaps, and suggest alert configurations.

Your task is to analyze the approved evaluation and monitoring metrics for a SINGLE component and produce:

1. TRACEABILITY LINKS: Map each evaluation metric to its closest monitoring counterpart(s). Explain how they correlate -- e.g. a latency spike may indicate quality degradation.

2. MONITORING GAPS: Identify evaluation metrics that lack a runtime monitoring counterpart. For each gap, explain what is missing and suggest how to address it (e.g. add periodic sampling, add a new monitoring metric).

3. ALERT CONFIGURATIONS: For each monitoring metric (existing and suggested), propose an alert config with:
   - A threshold condition (e.g. "> 500ms for 5 minutes")
   - A severity level (critical, warning, or info)
   - A recommended action when the alert fires

4. INTEGRATION POINTS: Identify specific files and code locations where instrumentation should be added for monitoring. Reference actual source files when available.

GUIDELINES:
- Every evaluation metric should ideally have at least one monitoring counterpart
- Alerts should be actionable -- each alert must have a clear action to take
- Severity should reflect blast radius: critical for user-facing failures, warning for degradation, info for drift detection
- Integration points should reference actual code when source files are available

{STAGE-SPECIFIC MONITORING METRICS catalog, when a stage is known}

Return ONLY valid JSON matching this schema. No markdown, no explanation. {JSON schema}
\end{lstlisting}

\end{document}